\documentclass{article}

\usepackage{main,times}

\usepackage{amsmath}
\usepackage{amsfonts}
\usepackage{bm}
\usepackage{amssymb}
\usepackage{mathtools}
\usepackage{amsthm}
\usepackage{dsfont}
\usepackage{bbm} 
\usepackage{stmaryrd} 
\usepackage{enumitem}
\usepackage{todonotes}

\usepackage{array}
\usepackage{pifont}
\usepackage{threeparttable}
\usepackage{pifont}
\usepackage{threeparttable}

\usepackage{colortbl}
\usepackage{mdframed}

\definecolor{rightblue}{RGB}{76,114,176} 
\definecolor{rightorange}{RGB}{221,132,82} 
\definecolor{aliceblue}{rgb}{0.94, 0.97, 1.0} 
\definecolor{darkcerulean}{rgb}{0.03, 0.27, 0.49} 
\definecolor{iris}{rgb}{0.35, 0.31, 0.81} 
\definecolor{carmine}{rgb}{0.59, 0.0, 0.09} 
\definecolor{green(munsell)}{rgb}{0.0, 0.66, 0.47} 
\definecolor{celadon}{rgb}{0.67, 0.88, 0.69} 
\definecolor{bluerow}{rgb}{0.0, 0.53, 0.74} 
\definecolor{lightorange}{RGB}{255, 219, 187} 
\definecolor{lavenderblue}{rgb}{0.8, 0.8, 1.0}
\definecolor{blue(pigment)}{rgb}{0.2, 0.2, 0.6}
\definecolor{blue-violet}{rgb}{0.54, 0.17, 0.89}

\definecolor{blueseaborn}{RGB}{1,115,178}
\definecolor{orangeseaborn}{RGB}{222,143,6}
\definecolor{greenseaborn}{RGB}{1,158,115}

\usepackage{acronym}

\usepackage{hyperref}
   \hypersetup{
     final=true,
     plainpages=false,
     pdfstartview=FitV,
     pdftoolbar=true,
     pdfmenubar=true,
     bookmarksopen=true,
     bookmarksnumbered=true,
     breaklinks=true,
     linktocpage=true,
     colorlinks=true,
     linkcolor=blue-violet, 
     urlcolor=iris, 
     citecolor=darkcerulean, 
     anchorcolor=black
}

\usepackage{url}            
\usepackage{nicefrac}       
\usepackage{microtype}      

\usepackage{algorithm}
\usepackage{algpseudocode}

\usepackage{caption} 
\usepackage{subcaption}

\usepackage{booktabs}
\usepackage{multirow}
\usepackage{graphicx}

\usepackage[capitalize]{cleveref}

\newtheorem{asm}{Assumption}[section]

\def\propcolor{lavenderblue!12}
\usepackage{mdframed}

\newmdtheoremenv[topline=false, bottomline=false, leftline=false, rightline=false, backgroundcolor=\propcolor,%
innertopmargin=\topskip, splittopskip=\topskip, skipbelow=\baselineskip, skipabove=\baselineskip]{boxprop}{Proposition}[section]

\newmdtheoremenv[topline=false, bottomline=false, leftline=false, rightline=false, backgroundcolor=\propcolor,%
innertopmargin=\topskip, splittopskip=\topskip, skipbelow=\baselineskip, skipabove=\baselineskip]{boxcor}{Corollary}[boxprop]

\newmdtheoremenv[topline=false, bottomline=false, leftline=false, rightline=false, backgroundcolor=\propcolor,%
innertopmargin=\topskip, splittopskip=\topskip, skipbelow=\baselineskip, skipabove=\baselineskip]{boxlem}{Lemma}[section]

\newmdtheoremenv[topline=false, bottomline=false, leftline=false, rightline=false, backgroundcolor=\propcolor,%
innertopmargin=\topskip, splittopskip=\topskip, skipbelow=\baselineskip, skipabove=\baselineskip]{boxdef}{Definition}[section]

\crefname{thm}{theorem}{theorems}
\Crefname{thm}{Theorem}{Theorems}

\crefname{lem}{lemma}{lemmas}
\Crefname{lem}{Lemma}{Lemmas}

\crefname{prop}{proposition}{propositions}
\Crefname{prop}{Proposition}{Propositions}

\crefname{defn}{definition}{definitions}
\Crefname{defn}{Definition}{Definitions}

\crefname{cor}{corollary}{corollaries}
\Crefname{cor}{Corollary}{Corollaries}

\crefname{asm}{assumption}{assumptions}
\Crefname{asm}{Assumption}{Assumptions}

\crefname{boxthm}{theorem}{theorems}
\Crefname{boxthm}{Theorem}{Theorems}

\crefname{boxlem}{lemma}{lemmas}
\Crefname{boxlem}{Lemma}{Lemmas}

\crefname{boxcor}{corollary}{corollaries}
\Crefname{boxcor}{Corollary}{Corollaries}

\crefname{boxprop}{proposition}{propositions}
\Crefname{boxprop}{Proposition}{Propositions}

\crefname{boxdef}{definition}{definitions}
\Crefname{boxdef}{Definition}{Definitions}

\makeatletter
\AddToHook{cmd/appendix/before}{\def\cref@section@alias{appendix}\def\cref@subsection@alias{appendix}}
\makeatother

\usepackage{tikz}
\usepackage{varwidth}

\usepackage{stmaryrd}

\usepackage{dirtytalk}
\MakeRobust{\say} 

\usepackage{makecell}

\usepackage{wrapfig}

\usepackage{enumitem}

\usepackage{nicematrix}

\usepackage{mathrsfs}

\usepackage{overpic}

\usepackage{lipsum}

\usepackage{soul}

\definecolor{mygreen}{HTML}{5a9027}

\definecolor{mygreen}{HTML}{5a9027}

\newcommand{\Dataset}{\mathcal{D}}
\newcommand{\identity}{\mathrm{I}}

\newcommand{\iid}{$\mathrm{iid}$\xspace}

\newcommand{\EE}{\mathbb{E}}
\newcommand{\NN}{\mathbb{N}}
\newcommand{\PP}{\mathbb{P}}

\newcommand{\RR}{\mathbb{R}}

\newcommand{\mrm}[1]{\mathrm{#1}}

\newcommand{\eg}{e.g. }

\newcommand{\vx}{{\bf x}}
\newcommand{\vy}{{\bf y}}

\newcommand{\RNum}[1]{\uppercase\expandafter{\romannumeral #1\relax}}

\usepackage{mleftright} 

\usepackage{xspace}

\newcommand{\KL}[2]{d_\mrm{KL}\mleft(#1||#2\mright)}
\newcommand{\Exp}[2]{\EE_{#1}\mleft[#2\mright]}
\newcommand{\abs}[1]{\lvert#1\rvert}

\DeclareMathOperator*{\argmax}{arg\,max}
\DeclareMathOperator*{\argmin}{arg\,min}

\allowdisplaybreaks

\title{From Data to Rewards: a Bilevel Optimization Perspective on Maximum Likelihood Estimation}

\iclrfinalcopy

\author{%
  Abdelhakim Benechehab\textsuperscript{ $\star$ 1,2}, 
  Gabriel Singer\textsuperscript{ $\star$ 1}, 
  Corentin Léger\textsuperscript{1}, 
  Youssef Attia El Hili\textsuperscript{1}, \\
  \textbf{Giuseppe Paolo\textsuperscript{3}, 
  Albert Thomas\textsuperscript{1},
  Maurizio Filippone\textsuperscript{4}, 
  Bal\'{a}zs K\'{e}gl\textsuperscript{1}} \\[0.5em]
  \textsuperscript{1} Huawei Noah's Ark Lab, Paris \\
  \textsuperscript{2} Department of Data Science, EURECOM \\
  \textsuperscript{3} Cognizant AI Lab, Paris \\
  \textsuperscript{4} Statistics Program, KAUST \\
  \fontsize{8}{8}\selectfont
  \tt $\star$ Equal contribution. $\quad$ Correspondence: \href{mailto:abdelhakim.benechehab@gmail.com}{abdelhakim.benechehab@gmail.com}
}

\begin{document}

\acrodef{MLE}{Maximum Likelihood Estimation}
\acrodef{PG}{Policy Gradient}
\acrodef{Bi-O}{Bilevel Optimization}

\addtocontents{toc}{\protect\setcounter{tocdepth}{0}}

\maketitle

\begin{abstract}
Generative models form the backbone of modern machine learning, underpinning state-of-the-art systems in text, vision, and multimodal applications. While Maximum Likelihood Estimation has traditionally served as the dominant training paradigm, recent work have highlighted its limitations, particularly in generalization and susceptibility to catastrophic forgetting compared to Reinforcement Learning techniques, such as Policy Gradient methods. However, these approaches depend on explicit reward signals, which are often unavailable in practice, leaving open the fundamental problem of how to align generative models when only high-quality datasets are accessible. In this work, we address this challenge via a Bilevel Optimization framework, where the reward function is treated as the optimization variable of an outer-level problem, while a policy gradient objective defines the inner-level. We then conduct a theoretical analysis of this optimization problem in a tractable setting and extract insights that, as we demonstrate, generalize to applications such as tabular classification and model-based reinforcement learning. We release the code at \href{https://github.com/abenechehab/nll_to_po}{https://github.com/abenechehab/nll\_to\_po}.
\end{abstract}

\section{Introduction}
\label{sec:intro}

Generative models have become central to modern machine learning research, driving advances in text~\citep{brown2020languagemodelsfewshotlearners, deepseekr1}, image~\citep{rombach2021highresolution, ramesh2021zero}, and multimodality~\citep{zhang2024,bai2025qwen25vl,fu2025vita,lajszczak2024,Yin2024} under the umbrella of “Generative AI” (\textit{GenAI}). Their ability to synthesize realistic content has made them foundational in applications ranging from decision making~\citep{shi2025,kim2024,intelligence2025} to scientific discovery~\citep{Manica2023,lu2024aiscientist}.

Traditionally, such models are trained via \ac{MLE}, where the parameters of the generative model are optimized to maximize the probability of observed data. This approach provides a principled framework for fitting models to large datasets and remains the backbone of many generative learning pipelines. Notably, this approach is omnipresent in today's \textit{Large Language Models} (LLMs) through the \textit{next token prediction} paradigm~\citep{vaswani2023attentionneed,brown2020languagemodelsfewshotlearners,deepseekr1}. 

However, recent breakthroughs in LLMs research, demonstrate the limitations of \ac{MLE} alone. Techniques based on \ac{PG} methods~\citep{bellman1958}, such as \textit{Reinforcement Learning from Human Feedback}~\citep{christiano2017, stiennon2020} and more recently \textit{Reinforcement Learning from Verifiable Rewards}~\citep{shao2024, deepseekr1}, have proven more effective than supervised fine-tuning at aligning models with human preferences and improving generation quality~\citep{shenfeld2025,lai2025,swamy2025}. These methods leverage explicit or implicit reward signals to guide training beyond likelihood objectives.

In many real-world scenarios, explicit reward functions for the tasks we aim to solve are not readily available. Instead, we often have access to high-quality datasets that we wish to use for aligning our models. Depending on the structure of these datasets, several techniques have been proposed to derive reward functions, such as from preference data~\citep{rafailov2023direct} or from demonstrations~\citep{finn2016connection,finn2016guided} when framed within a Markov Decision Process (MDP) formalism. Despite these advances, the fundamental question surrounding this problem remains unresolved:
\begin{center}
\emph{Can we learn an \textbf{implicit reward function} from \textbf{unlabeled data}, and exploit the well-developed \textbf{policy optimization} literature to train models \textbf{more effectively} than with \ac{MLE}?}
\end{center}

In this paper, we propose the following contributions toward addressing this question:
\begin{itemize}
\item \textbf{Bilevel optimization perspective on \ac{MLE}:} We reinterpret the \ac{MLE} training objective as a \ac{Bi-O} problem, where the outer-level problem optimizes over the reward function, while the inner-level problem is defined by a \ac{PG} objective with respect to the model parameters.
\item \textbf{Theoretical analysis:} We study this formulation under a Gaussian data distribution with the reward given by a negatively scaled distance in the output space, deriving insights into the theoretically optimal parameters of the reward function.
\item \textbf{Practical algorithms:} Guided by the theoretical analysis and leveraging implicit differentiation solvers, we propose two practical algorithms for addressing the bilevel optimization problem. We evaluate these algorithms on two \ac{MLE} applications: tabular classification and model-based reinforcement learning.
\end{itemize}

The remainder of the paper is organized as follows. \cref{sec:related} situates our work within the relevant literature, and \cref{sec:prelims} introduces the problem setup and motivates our approach. In \cref{sec:solve_tractable}, we address the bilevel optimization problem in the Gaussian case, while \cref{sec:solve_general} considers the general setting using implicit differentiation. We then present experimental results in \cref{sec:applications} and conclude with a discussion in \cref{sec:conclusion}.

\section{Related Work}
\label{sec:related}

\paragraph{\ac{PG} vs \ac{MLE} for Generative models.} Generative models aim to capture the underlying distribution of observed data, with the goal of synthesizing realistic samples afterwards, \eg text generation~\citep{brown2020languagemodelsfewshotlearners} and image generation~\citep{rombach2021highresolution, ramesh2021zero}. 
Many of the existing generative modeling approaches such as Autoregressive models~\citep{Radford2018ImprovingLU,vaswani2023attentionneed,Radford2019LanguageMA}, Variational AutoEncoders~\citep{Kingma2013AutoEncodingVB,higgins2017betavae}, Generative Adversarial Networks~\citep{goodfellow2014, arjovsky2017}, Diffusion Models~\citep{sohldickstein2015, rombach2021highresolution}, can be framed through the lens of \ac{MLE} or its approximations. 
However, and especially in the context of sequence generation, \ac{MLE} in autoregressive models has been proven to suffer from compounding errors and exposure bias, among other problems~\citep{tan2019connectingdotsmlerl, bahdanau2017, ranzato2016, bengio2015sch, venkatraman2015, benechehab2024}. 
As an alternative approach, \ac{PG} methods have emerged as a more effective way to sample the output space when a reward function is available~\citep{bahdanau2017}. 
Beyond vanilla \ac{PG}, more sophisticated methods have been developed, such as Reward-Augmented Maximum Likelihood~\citep{norouzi2016, volkovs2011}, where a reward-based stationary sampling distribution is defined, Softmax Policy Gradient~\citep{Ding2017ColdStartRL}, an intermediate approach between sampling the model and sampling a reward-based distribution, and MIXER~\citep{ranzato2016}, a scheduling approach that gradually transitions from \ac{MLE} to \ac{PG} using the REINFORCE algorithm~\citep{Williams1992}. 
Besides autoregressive models, policy gradient methods have also been used to train (or finetune) Diffusion models~\citep{black2024training, uehara2024, zekri2025}, and GANs~\citep{Biswajit2017, yu2017seqgan}. 

\paragraph{Reward models.} Policy Gradient methods constitute one class of algorithms for solving MDPs~\citep{bellman1958}, the central formalism underpinning the RL field. Training generative models with \ac{PG} methods builds on the formulation of the task as an MDP.
In this setting, the reward function plays a pivotal role.
The most direct way of learning a reward model is via supervised learning from past interactions, as done in \textit{Model-based Reinforcement Learning}~\citep{Chua2018,Janner2019,Yu2020,Hafner2021,Kegl2021,benechehab2025zeroshot}. 
Beyond the supervised approach, several other paradigms for reward learning have been developed. 
\textit{Learning from Demonstrations} includes Inverse RL methods~\citep{abbeel2004,ziebart2008,finn2016connection,finn2016guided} that learn a reward model $R_\theta$ under which demonstrations of the form $(s,a,s_{\text{next}})$ are optimal. 
Another paradigm, \textit{Learning from Goals}, defines the reward function with respect to reaching a goal $g$ in the state space $\mathcal{S}$~\citep{liu2022goal}. 
In this setting, goal attainment has been modeled in terms of spatial distances~\citep{nachum2018,mazzaglia2024}, temporal distances~\citep{Hartikainen2020Dynamical,wang2025founder}, and semantic similarity~\citep{sontakke2023roboclip,fan2022minedojo}. 
The \textit{Learning from Preferences} approach relies on transforming preference data of the form $(\tau_0 \succ \tau_1)$, where $\tau_i$ is a trajectory $(s_1,a_1,\hdots,s_{\mid \tau_i \mid},a_{\mid \tau_i \mid})$ and $\succ$ is a preference relationship, into a reward model using the \textit{Bradley-Terry} model~\citep{bradley_terry1952}. 
Reward models learned from preference data have enabled significant progress in \textit{post-training} generative models~\citep{kim2023preference,touvron2023llama2,rafailov2023direct,song2024preference}. 
Starting with \textit{InstructGPT}~\citep{ouyang2022}, this approach has become a standard for improving targeted aspects of LLMs, \eg safety~\citep{dai2024safe}, as well as for applications such as mathematical reasoning~\citep{xin2025deepseekproverv,shao2024,luong2024} and code generation~\citep{deepseekr1}. 

\paragraph{Bilevel optimization.} 

\ac{Bi-O} was originally introduced in economics and game theory by \cite{stackelberg1934} to model hierarchical decision-making problems between a leader and a follower.
More broadly, \ac{Bi-O} offers a framework for addressing problems with hierarchical structures, where the task is to optimize two interdependent objective functions: an \emph{inner-level} objective and an \emph{outer-level} objective.
In machine learning, \ac{Bi-O} was first applied to feature selection~\citep{bennett2006} and was later extended to a wide spectrum of applications, including hyperparameter optimization~\citep{mackay2018,franceschi2017,pedregosa2016}, reinforcement learning~\citep{hong2022,nikishin2021}, and meta-learning~\citep{franceschi2018}. 
Various \ac{Bi-O} solvers have been proposed to address different regularity conditions on the inner- and outer-level objectives. Among these, \textit{automatic differentiation}-based approaches compute gradients of the outer-level objective by differentiating through the iterative steps of the inner-level optimization algorithm~\citep{wengert1964,linnainmaa1976,domke2012,franceschi2017}. 
In parallel, \textit{implicit differentiation} methods~\citep{bengio2000} leverage the implicit differentiation theorem to approximately estimate the gradient of the outer loss by solving a linear system~\citep{pedregosa2016,chen2021closing,ji2021,arbel2022amortized}. Beyond alternating methods, \cite{dagreou2024} introduce a framework where inner- and outer-level variables evolve jointly within a single training loop. 
\ac{Bi-O} has also been generalized to functional settings~\citep{petrulionyte2024}, where the inner-level optimization is carried out over functions in infinite-dimensional spaces. 
In the context of generative models, some approaches enhance the training efficiency of energy-based latent variable models through bilevel formulations~\citep{bao2020,kan2022}, while \cite{xiao2025a} propose a bilevel framework for tuning hyperparameters and noise schedules in diffusion models.

\paragraph{Bilevel Reinforcement Learning.} 

Bilevel RL optimizes an outer-level objective in the reward parameters, often a policy alignment signal, while an inner loop learns a policy under that reward~\citep{gaur2025, shen2024, yang2025}. This framework has been applied in areas such as reward shaping \citep{zou2019reward} and RLHF \citep{chakraborty2024}. The closest work to ours is \citep{zeng2022maximum}, which combines \ac{MLE} with the \textit{Maximum-Entropy} inverse RL framework. However, they focus on control tasks in the episodic RL setting while we aim at providing a general framework for any data modality and any \ac{MLE} task.

\section{Preliminaries}
\label{sec:prelims}

In \cref{subsec:motivation}, we begin by motivating the idea of learning reward functions from data, outlining scenarios in which \ac{PG} methods may be preferred over \ac{MLE}. We then formally define the problem setup in \cref{subsec:problem}.


\subsection{Motivation}
\label{subsec:motivation}

In Reinforcement Learning, \ac{PG} methods are traditionally viewed as producing unbiased yet high-variance gradient estimates, especially in long-horizon or high-dimensional tasks~\citep{greensmith2001}. In contrast, \ac{MLE} has historically served as the dominant paradigm in supervised learning and probabilistic modeling~\citep{Akaike73}. 
However, in the current era of large pretrained models and advanced RL algorithms, these limitations have become less restrictive, giving rise to many cases where \ac{PG} methods are more advantageous than \ac{MLE}.

A first phenomenon is the \textit{mismatch between training objectives and evaluation metrics}. In sequence prediction, for instance, evaluation scores such as BLEU or ROUGE (widely used for machine translation) do not decompose into token-level likelihoods. While for the widely used autoregressive models \ac{MLE} is restricted to maximizing token-level likelihoods, \ac{PG} methods directly optimize sequence-level rewards and naturally account for this discrepancy~\citep{norouzi2016,Ding2017ColdStartRL,ranzato2016}. 


Another key phenomenon is \textit{catastrophic forgetting}. When adapting large language models to downstream tasks through post-training, it is often desirable to preserve prior knowledge while specializing to new distributions. Recent studies~\citep{shenfeld2025,lai2025,swamy2025} suggest that on-policy RL fine-tuning achieves this balance more effectively than supervised fine-tuning, since its updates converge to solutions closest in KL divergence to the original policy.


Taken together, these observations motivate our approach: rather than maximizing the likelihood directly, we propose a general framework that interprets data signals as reward functions, thereby also enabling \ac{PG} optimization.


\subsection{Problem setup}
\label{subsec:problem}

Let $(\Omega, \mathcal{F}, \PP)$ be a probability space, and let $X:\Omega \to \mathcal{X}$ and $Y:\Omega \to \mathcal{Y}$ be two random variables, with $\mathcal{X} \subseteq \RR^{m}$ and $\mathcal{Y} \subseteq \RR^n$, where $(n,m)\in \NN_{\star}^{2}$. Consider a maximum likelihood estimation problem where we observe $N$ \iid realizations $\Dataset = \{(\vx_i, \vy_i)\}_{i=0}^N$ from a fixed unknown distribution over $\mathcal{X} \times \mathcal{Y}$. The goal is to model the conditional distribution $Y|X \sim q$ using a parametric model $\widehat{Y}|X \sim \hat{p}_\theta$ where $\theta \in \Theta := \RR^{d_\theta}$ are parameters spanning a finite dimensional space with dimension $d_\theta$. In the \ac{MLE} formalism, we optimize the parameters $\theta$ by maximizing the log-likelihood, equivalently seen as a Kullback-Leibler divergence minimization~\citep{Akaike73} (denoted as $d_\mrm{KL}$):

\begin{equation}
\label{eq:mle}
\tag{MLE}
\theta^\star = \argmin_{\theta \in \Theta} \Exp{X}{\KL{q\mleft(\cdot|X\mright)}{\hat{p}_\theta \mleft(\cdot|X\mright)}} = \argmax_{\theta \in \Theta} \mathbb{E}_{X}\Exp{Y\mid X \sim q}{\log{\hat{p}_\theta \mleft(Y|X\mright)}}
\end{equation}

A parallel approach, based on reinforcement learning, consists in maximizing a reward function $r:\mathcal{Y} \times \mathcal{Y} \to \RR$ that evaluates the quality of generated $\hat{\vy}$ against the true observations $\vy$, resulting in the Policy Gradient (PG) objective. 
Here we state the entropy-regularized PG objective, a variant that is commonly considered in RL algorithms~\citep{haarnoja2017reinforcement, haarnoja2018sac, wen2024entropy}:

\begin{equation}
\label{eq:pg}
\tag{PG}
\theta^\star = \argmax_{\theta \in \Theta} \mathbb{E}_{X}\Exp{Y\mid X \sim q}{\Exp{\widehat{Y} \mid X \sim \hat{p}_{\theta}}{r(\widehat{Y},Y)} + \lambda \mrm{H}\mleft( \hat{p}_\theta \mright)},
\end{equation}

where $\lambda > 0$ is a parameter controlling the strength of the regularization, and $\mrm{H}$ denotes the entropy.

In this work, we ask whether the reward function itself can be seen as an optimization variable $r$ over a Hilbert space $\mathcal{H}$. 
The \emph{optimal reward function} is then determined based on the \ref{eq:mle} objective, which now represents the outer-level of the following bilevel optimization problem:

\begin{boxdef}[Bilevel Optimization problem]
\label{def:bilevel_prob}
\small
\begin{equation}
\label{eq:bo}
\tag{Bi-O}
\max_{r \in \mathcal{H}} \mathbb{E}_{X}\Exp{Y\mid X \sim q}{\log{\hat{p}_{\theta^{\star}_r} \mleft(Y|X\mright)}} \quad \text{s.t.} \quad \theta^{\star}_r = \argmax_{\theta \in \Theta} \mathbb{E}_{X}\Exp{Y\mid X \sim q}{\Exp{\widehat{Y} \mid X \sim \hat{p}_{\theta}}{r(Y',Y)}+ \lambda \mrm{H}\mleft( \hat{p}_\theta \mright)}
\end{equation}
\end{boxdef}


\section{Solving \ref{eq:bo} in a tractable case}
\label{sec:solve_tractable}

The objective of this section is to analyze the bilevel optimization problem \ref{eq:bo} under specific assumptions on the data-generating distribution and the reward parametrization, in which both the inner- and outer-level problems admit closed-form solutions.

\subsection{Theoretical analysis}
\label{subsec:theory}

We start our analysis by stating the following assumptions, which will prove useful in the establishment of our main results:

\begin{asm}[Gaussian density model]
\label{ass:model_general}
\small
We assume that both the true conditional density $q$ and the model density $\hat{p}_\theta$ are Gaussian distributions with linear mean functions and fixed covariance matrices:
\[
Y \mid X \sim q := \mathcal{N}(\Lambda X, \Sigma), 
\quad 
\widehat{Y} \mid X \sim \hat{p}_{\theta} := \mathcal{N}(\mrm{A} X, \mrm{B}),
\]
where $\Lambda \in \RR^{n \times n}$,  $\Sigma \in \mrm{S}^{++}_{n}(\RR)$\footnote{We denote by $\mrm{S}^{++}_{n}(\mathbb{R})$ the set of real symmetric
positive definite $n\times n$ matrices.
}, and $\theta := (\mrm{A}, \mrm{B}) \in \Theta := \RR^{n \times n} \times \mrm{S}^{++}_{n}(\RR)$.
\end{asm}

\begin{asm}[Reward model]
\label{ass:reward}
\small
Let $\mrm{U} \in S^{++}_n(\mathbb{R})$, we define the reward model as the following quadratic form:
$
\forall (\widehat{Y},Y)\in \mathbb{R}^{n}\times \mathbb{R}^{n},\quad r_\mrm{U}(\widehat{Y},Y) = -(\widehat{Y} - Y)^T \mrm{U} (\widehat{Y} - Y).$
\end{asm}

We first notice that this choice of parametrization is valid as the resulting reward function is maximized in $Y$. Furthermore, this parametrization enables that for any $\mrm{U} \in \mathbb{R}^{n\times n}$, $r_\mrm{U}$ is an element of the Hilbert space $\mathcal{H}$ of square-integrable real-valued functions with a weighted measure.
We refer the interested reader to \cref{app:hilbert} for a technical definition of $\mathcal{H}$ and a proof of this statement. We now state the main results, showcasing closed-form solutions of the \ref{eq:bo} problem under the previous assumptions. 

\begin{boxprop}[Closed-form solution]
\label{prop:outer}
\small
Under \cref{ass:model_general} and \ref{ass:reward}, the \ref{eq:bo} problem
\begin{align*}
\mrm{U}^{\star} &= \argmax_{U\in S_{n}^{++}(\RR)} \mathbb{E}_{X}\Exp{Y\mid X \sim q}{\log{\hat{p}_{\theta^{\star}_\mrm{U}} (Y|X)}} \\
\text{s.t.} \quad \theta^\star_\mrm{U} &= \argmax_{\theta \in \Theta} \mathbb{E}_{X}\Exp{Y\mid X \sim q}{
\Exp{\widehat{Y} \mid X \sim \hat{p}_{\theta}}{
-(\widehat{Y} - Y)^T \mrm{U} (\widehat{Y} - Y)} + \lambda \mrm{H}(\hat{p}_{\theta})}
\end{align*}
has exactly one solution that writes:
\begin{center}
$
\mrm{U}^{\star}=\frac{\lambda}{2}\Sigma^{-1}.$
\end{center}
\end{boxprop}
The proof of \cref{prop:outer} is deferred to \cref{app:inner}. 


\begin{boxcor}[Isotropic case]
\label{cor:isotropic}
\small
For $\mrm{B} = \sigma^2 \identity_n$, the set of solutions of \ref{eq:bo} is:
\begin{equation*}
    \mrm{U}^\star \in \mrm{F}_{\lambda,\Sigma}:=\left\{ \mrm{U} \in S_n^{++}(\mathbb{R}) \, \middle| \, \operatorname{Tr}(\mrm{U})= \frac{\lambda n^2}{2\operatorname{Tr}(\Sigma)} \right\}.
\end{equation*}
\end{boxcor}

Note that, for any given $\lambda>0,\Sigma \in \mrm{S}_{n}^{++}(\RR),$ $\mrm{F}_{\lambda,\Sigma}\neq \varnothing$ since $\frac{\lambda n}{2\operatorname{Tr}(\Sigma)}\identity_n \in \mrm{F}_{\lambda,\Sigma}$ which corresponds to reward functions we consider for the empirical experiments in \cref{subsec:empirical}.

\paragraph{Interpretation as Mahalanobis distance.} We observe that the optimal reward function is characterized by a matrix $\mrm{U}^\star \in \mrm{S}^{++}_{n}(\mathbb{R})$ that is inversely proportional to the covariance matrix of the data-generating process. With this choice, the reward admits a clear interpretation as the negative squared Mahalanobis distance~\citep{mahalanobis1936}, which measures the distance between the sample $\widehat{Y}$ and the Gaussian distribution centered at $Y$ with covariance matrix $\Sigma$. 
This perspective suggests that, in practice, the noisier the data, the less strongly the model should be penalized for deviations from samples. Finally, the scaling by $\lambda$ serves to balance the two competing optimization objectives, reward maximization and entropy regularization.

\paragraph{Interpretation as reverse KL minimization.} An interesting observation arises when substituting the optimal reward parametrization $\mrm{U}^\star$ into the inner-level objective: the \ac{PG} formulation becomes equivalent to minimizing the reverse KL divergence between the model distribution $\hat{p}_\theta$ and the data-generating distribution $q$. This connection provides an explanation for the empirical results presented in the next section. We therefore state it formally as a corollary, with the proof deferred to \cref{app:corr45}:


\begin{figure}[t] 
 \centering
 \includegraphics[width=0.85\textwidth]{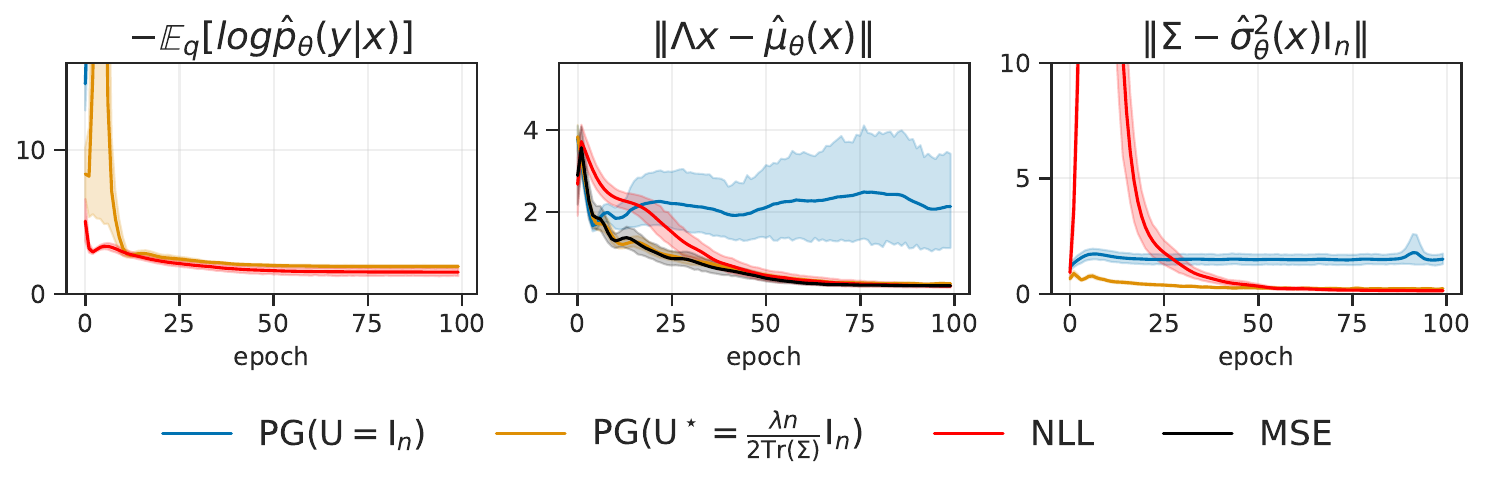}
 \caption{\textbf{Synthetic data experiment.} The \ac{PG} loss when paired with the optimal reward function matches the NLL-trained baseline in terms of NLL (left panel), all while having faster convergence in terms of moment matching (center and right panels for the mean and variance, respectively).}
 \label{fig:toy}
\end{figure}

\begin{boxcor}
\label{cor:reverse_kl}
\small
Under the assumptions of \cref{prop:outer}, the optimal parameters $\theta^\star_{\mathrm{U}^\star}$ obtained from the lower-level problem with $\mathrm{U}^\star = \frac{\lambda}{2}\Sigma^{-1}$ minimize the reverse KL divergence between  $\hat{p}_\theta$ and $q$:
\begin{center}
$\theta^\star_{\mathrm{U}^\star} = \argmin_{\theta \in \Theta} \mathbb{E}_{X} \left[ d_{KL} \left( \hat{p}_\theta(\cdot|X) \,\|\, q(\cdot|X) \right) \right].$
\end{center}
\end{boxcor}


\subsection{Empirical validation}
\label{subsec:empirical}

In this section, we empirically evaluate the theoretical results from \cref{subsec:theory}. 
To this end, we generate synthetic data that satisfy \cref{ass:model_general}: $\Dataset = \{(\vx_i, \vy_i)\}_{i=0}^N$, where $\vx_i \sim \mathcal{U}([-5,5]^{n})$ ($\mathcal{U}$ denotes the uniform distribution), and $\vy_i \sim q(.|\vx_i) := \mathcal{N}(\Lambda \vx_i, \Sigma)$ with diagonal covariance matrix $\Sigma=\beta^2 \identity_n$ and $\beta > 0$. 
For the model $\hat{p}_\theta$, we relax the linearity and homoscedasticity assumptions by considering a neural network that parametrizes a Gaussian distribution, in which both the mean function and the diagonal covariance matrix depend on the input: $\hat{p}_\theta(.|\vx_i) := \mathcal{N}(\mu_\theta(\vx_i), \sigma^2_\theta(\vx_i) \identity_n)$. 
We compare baselines trained with the \textit{negative log-likelihood} (NLL) and \textit{mean squared error} (MSE) losses against \ac{PG} variants, using either a negative squared distance reward $\mrm{U}=\identity_n$ or the optimal reward function derived in \cref{cor:isotropic} with $\mrm{U}^\star=\frac{\lambda n}{2\operatorname{Tr}(\Sigma)}\identity_n$.


\cref{fig:toy} shows how the validation NLL and moment-matching errors (mean and covariance) change over training.
Consistent with theory, we observe that adjusting the reward function with the optimal matrix $\mrm{U}^\star$ yields a learning curve nearly identical to the NLL baseline (yellow and red curves in the left panel of \cref{fig:toy}). 
Moreover, the \ac{PG} variant with the optimal matrix converges faster than the NLL baseline in matching the moments of the data-generating distribution (center and right panel). 
Finally, we note that the vanilla \ac{PG} method (with $\mrm{U}=\identity_n$) suffers from a diverging NLL due to the variance shrinking to zero for some values of $\lambda$ which leads to numerical instabilities.

\begin{figure}[h!]
 \centering
 \includegraphics[width=0.85\textwidth]{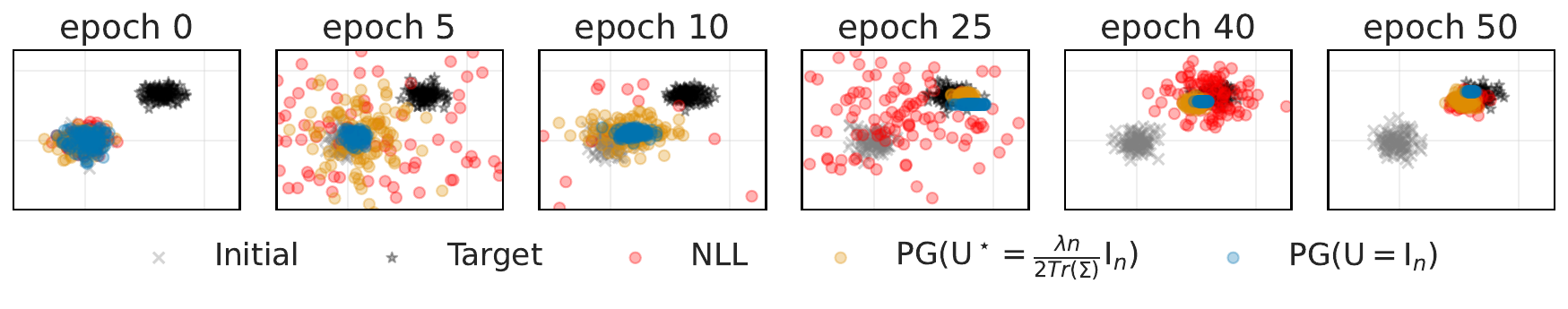}
 \caption{\textbf{Learned distributions comparison on  a single data point.} The \ac{PG} loss paired with the optimal reward function in \cref{cor:isotropic} shows optimal convergence, even when compared with the baseline directly optimizing the NLL.}
 \label{fig:dist}
\end{figure}

To gain further insight, \cref{fig:dist} shows the evolution of the learned distributions for a single training data point. 
In this illustrative example, the \ac{PG} variant with the optimal reward displays the most natural behavior in fitting the target distribution, unlike the NLL baseline, which initially causes the variance to increase sharply before reducing it to match the target variance. This behavior can be explained by \cref{cor:reverse_kl} since minimizing \( d_{KL} \left( \hat{p}_\theta \,\|\, q \right) \) is known to induce mode seeking behavior, thus encouraging \( \hat{p}_\theta \) to concentrate directly on the mode of \( q \).



\section{Solving \ref{eq:bo} in general}
\label{sec:solve_general}

In contrast to the previous section, where we assumed access to the data-generating distribution and provided a closed-form solution to problem \ref{eq:bo}, real-world applications typically do not satisfy such assumptions. Consequently, solving the bilevel optimization problem \ref{eq:bo} by directly optimizing the outer objective offers a more general approach applicable to a broader class of problems.

Bilevel optimization solvers can generally be divided into three categories. Explicit gradient methods treat the gradient update as a differentiable mapping and backpropagate through the unrolled optimization path of the inner-level problem~\citep{franceschi2017}. 
Gradient-free methods instead rely on evolutionary strategies, optimizing the outer objective while considering the inner problem as a black-box function~\citep{Song2020,feng2021}. Finally, implicit differentiation methods leverage the implicit function theorem to reformulate gradient estimation as the solution to a linear system~\citep{dagreou2024, petrulionyte2024}.

In this work, we focus on implicit differentiation, as explicit gradient methods often encounter memory issues from storing long computational graphs, while gradient-free approaches are generally limited by the curse of dimensionality.

\subsection{Implicit differentiation}
\label{subsec:implicit_diff}

Consider a reward parametrization $r_\phi$ with $\phi \in \Phi := \RR^{d_\phi}$, where $d_\phi$ denotes the dimension of the reward parameter space. The optimization of the outer-level problem can thus be restricted to the Hilbert space of reward functions spanned by parameters $\phi \in \Phi$ (see the appendix for an explicit construction in the case of the Mahalanobis parametrization). Within this setup, implicit differentiation treats the solution of the inner problem, $\theta^\star$, as an implicit function of $\phi$ and allows one to compute the best-response derivatives $\nabla_\phi \theta^\star(\phi)$ analytically via the implicit function theorem.

To proceed, we define an operator $\mrm{f}: \Phi \times \Theta \to \Theta := \RR^{d_\theta}$ whose roots characterize the inner-level optimal solution $\theta^\star(\phi)$. That is, for all $\phi \in \Phi$, we have $\mrm{f}(\phi, \theta^\star(\phi)) = 0$. Leveraging this property, the derivative of interest $\nabla_\phi \theta^\star(\phi)$ can be determined by solving for $\nabla_\phi \mrm{f}(\phi, \theta^\star(\phi))=0$:
\begin{equation}
\label{eq:implicit}
 \forall \phi \in \Phi, \quad \nabla_\theta f (\phi,\theta^\star(\phi)) \,\nabla_\phi \theta^\star(\phi) 
+ \nabla_\phi f (\phi,\theta^\star(\phi)) = 0,
\end{equation}
where $\nabla_\phi \theta^\star(\phi)$ is obtained by solving the linear system in Eq.~\eqref{eq:implicit}, enabling gradient descent on the outer problem via the chain rule.

In our bilevel optimization formulation, the operator $\mrm{f}$ arises naturally from the fixed-point characterization of the gradient update: $\mrm{f}(\phi,\theta) = \theta + \alpha \nabla_\theta \mathcal{L}_{\text{in}}(\phi,\theta) - \theta = \alpha \nabla_\theta \mathcal{L}_{\text{in}}(\phi,\theta)$ where $\mathcal{L}_{\text{in}}$ is the inner-level objective and $\alpha$ is a learning rate. Under this definition, the first-order optimality condition holds whenever the inner-level optimization converges to a local minimum $\theta^\star(\phi)$, where the gradient vanishes, which we assume is a plausible hypothesis given any modern stochastic optimizer (\eg Adam~\citep{kingma2017}).

\subsection{Empirical validation}
\label{subsec:empirical_implicit_diff}

In practice, we use TorchOpt~\citep{JMLR:TorchOpt}, a python package that enables differentiable optimization solvers that can be integrated with Pytorch~\citep{pytorch} neural network implementations. Precisely, we run the implicit differentiation-based solvers using the Conjugate Gradient algorithm for the linear system resolution, as in \citep{rajeswaran2019metalearningimplicitgradients}. We now compare the obtained results, in the same setup as \cref{subsec:empirical}, to get insights into the effectiveness of this kind of bilevel optimization solvers against MLP-based policies.

\begin{figure}[t]
 \centering
 \includegraphics[width=0.99\textwidth]{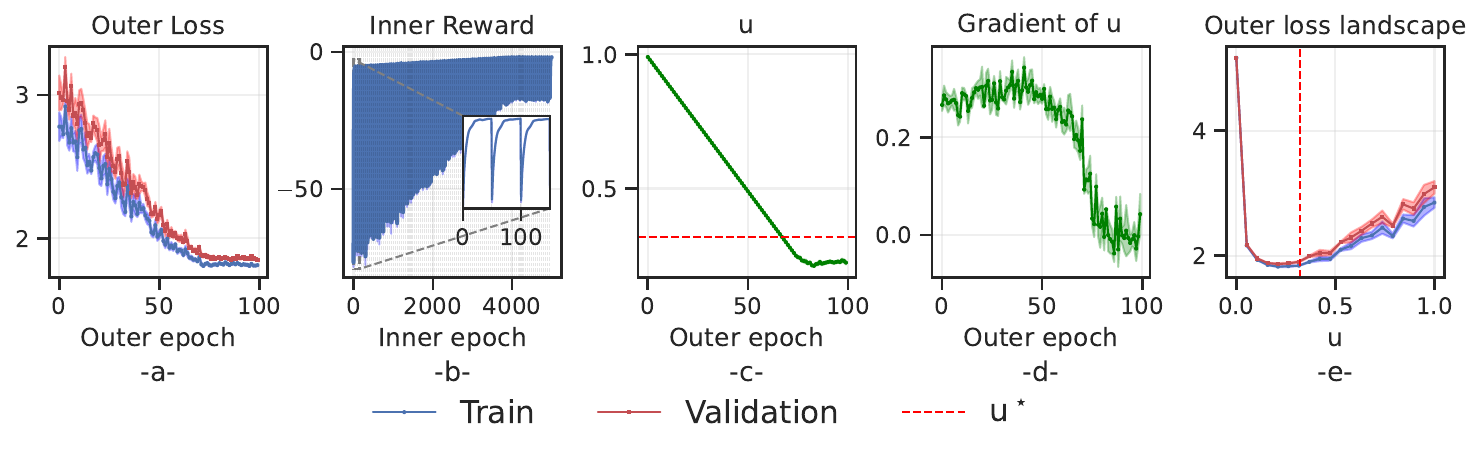}
 \caption{\textbf{Implicit differentiation solver on synthetic data experiment.} From left to right: \textbf{-a-} outer loss (NLL), \textbf{-b-} inner reward optimization loop, \textbf{-c-} trajectory of the reward parameter $\mrm{u}$, \textbf{-d-} gradient of the outer loss with respect to $\mrm{u}$, \textbf{-e-} the outer loss landscape.}
 \label{fig:implicit}
\end{figure}

\cref{fig:implicit} presents the results of running an implicit differentiation solver for 100 outer iterations, each with 50 inner iterations, and a learning rate of $10^{-2}$ for both optimization loops. The outer optimization variable is a single parameter (panel \textbf{-c-}) initialized at 1, which defines the diagonal Mahalanobis matrix for the reward: $\mrm{u}>0$ s.t. $\mrm{U} = \mrm{u} \cdot \identity_n$. Panel \textbf{-a-} illustrates the evolution of the outer loss (NLL evaluated on the optimal policy from the inner \ac{PG} loop), showing clear improvement relative to the initialization at 1 (which corresponds to the Euclidean distance). Additionally, the optimization parameter $\mrm{u}$ converges to a value close to the theoretical optimum $\mrm{u}^\star=\frac{\lambda n}{2\operatorname{Tr}(\Sigma)}$, as derived in \cref{cor:isotropic}. This convergence is further supported by panel \textbf{-e-}, which plots the outer loss landscape as a function of the reward parameter $\mrm{u}$, revealing a roughly convex landscape with a global minimum near the theoretical optimum. These results validate our intuition from the tractable case discussed in \cref{sec:solve_tractable}, even in the more general setting of MLP-based policies and stochastic optimization solvers within the implicit differentiation framework.

\section{Applications}
\label{sec:applications}

The goal of this section is to use intuition gained from the previous analysis to derive practical algorithms that we can validate on common \ac{MLE} tasks from the literature. 



\begin{figure}[h]
    \centering
    \begin{minipage}{0.48\textwidth}
        \begin{algorithm}[H]
            \caption{\textbf{PG}$(\mathrm{U}_{\text{he}}^\star)$ - heuristic}
            \label{alg:pg1}
            \textbf{Input:} Data $\Dataset = \{(\vx_i, \vy_i)\}_{i=0}^N$, model $\hat{p}_\theta$, $\lambda$  \\
            \textbf{1.} Estimate cov matrix $\hat{\Sigma} = \text{cov}(\{(\vy_i)\}_{i=0}^N)$ \; \\
            \textbf{2.} $\text{loss} \leftarrow \text{PG}(\mathrm{U}^{\star}_{\text{he}}(\lambda, \hat{\Sigma}))$ \; \\
            \textbf{3.} $\texttt{train\_policy}(\hat{p}_\theta, \Dataset, \text{loss})$ \; \\
            \textbf{Return:} learned model $\hat{p}_{\theta^\star}$
        \end{algorithm}
    \end{minipage}
    \hfill
    \begin{minipage}{0.48\textwidth}
        \begin{algorithm}[H]
            \caption{\textbf{PG}$(\mathrm{U}_{\text{im}}^\star)$ - implicit differentiation}
            \label{alg:pg2}
            \textbf{Input:} Data $\Dataset = \{(\vx_i, \vy_i)\}_{i=0}^N$, model $\hat{p}_\theta$, $\lambda$  \\
            \textbf{1.} $\mathrm{U}_{\text{im}}^\star \leftarrow \texttt{imp\_diff\_solver}(\Dataset,\lambda)$ \; \\
            \textbf{2.} $\text{loss} \leftarrow \text{PG}(\mathrm{U}_{\text{im}}^{\star})$ \; \\
            \textbf{3.} $\texttt{train\_policy}(\hat{p}_\theta, \Dataset, \text{loss})$ \; \\
            \textbf{Return:} learned model $\hat{p}_{\theta^\star}$
        \end{algorithm}
    \end{minipage}
\end{figure}

We build on the theoretical analysis in \cref{subsec:theory} to suggest a realistic way to estimate the optimal reward parametrization $\mrm{U}^\star$. The main challenge with this approach lies in estimating the covariance matrix-dependent term. As stated in \cref{alg:pg1}, we propose an approximate approach that estimates an empirical covariance matrix $\hat{\Sigma}$ from the training data. 
In parallel, we use the implicit differentiation-based bilevel solver to provide a gradient-based approach (\cref{alg:pg2}). 
Such an approach, is more general as it's not sensitive to the estimation error on the covariance matrix, nor requires the validity of the assumptions under which we derive our theoretical results.

In the following, we use both \cref{alg:pg1} and \cref{alg:pg2} to benchmark our method against vanilla \ac{PG} and NLL losses in two real-world applications: tabular classification, and model-based reinforcement learning.
Note that, in the experiments, we are effectively solving the inner-level problem of the \ref{eq:bo} formulation, while substituting the reward function either with the optimal matrices $\mathrm{U}_{\text{he}}^\star$ and $\mathrm{U}_{\text{im}}^\star$, or with the identity $\identity_n$ for the squared-distance baseline. 

\begin{wraptable}[10]{r}{0.35\textwidth}
\vskip -0.7cm
\centering
\resizebox{0.35\textwidth}{!}{
\begin{tabular}{l l c}
\toprule
\textbf{Dataset} & \textbf{Method} & \textbf{Accuracy}\tiny{$/ 10^{-2}\uparrow$} \\
\midrule
\multirow{3}{*}{\textit{Credit default}}
  & \textbf{NLL}                & $79.8 \,\pm\, .012$ \\ 
  & \textbf{PG}$(\identity_n)$  & $75.5 \,\pm\, .023$ \\ 
  & \textbf{PG}$(\mathrm{U}_{\text{he}}^\star)$ & $\mathbf{82.0} \,\pm\, .001$ \\ 
\midrule
\multirow{3}{*}{\textit{Poker}}
  & \textbf{NLL}                & $48.6 \,\pm\, .001$ \\ 
  & \textbf{PG}$(\identity_n)$  & $38.2 \,\pm\, .039$ \\ 
  & \textbf{PG}$(\mathrm{U}_{\text{he}}^\star)$ & $\mathbf{52.4} \,\pm\, .001$ \\ 
\bottomrule
\end{tabular}
}
\caption{\textbf{Accuracy.}}
\label{tab:uci_accuracy}
\end{wraptable}

\subsection{Tabular classification}
\label{subsec:tabular}

We evaluate our framework on two tabular classification datasets from the UCI repository~\citep{uci}. Specifically, we train a multiclass logistic regression model with the \ac{PG} loss, where the reward is defined using the distance between the one-hot encoded ground-truth labels and generations sampled from the model's softmax distribution.
We consider both multiclass (\textit{Poker}) and imbalanced binary classification (\textit{Credit default}). 
In the case of unbalanced datasets, accuracy alone can be misleading, in which case we additionally report the Area Under the ROC curve (AUC).

\begin{wraptable}[4]{r}{0.35\textwidth}
\vskip -2.0cm
\centering
\resizebox{0.25\textwidth}{!}{
\begin{tabular}{l c}
\toprule
\textbf{Method} & \textbf{AUC}\tiny{$/ 10^{-2}\uparrow$} \\
\midrule
\textbf{NLL}                       & $70.5 \,\pm\, .005$ \\ 
\textbf{PG}$(\identity_n)$         & $57.7 \,\pm\, .632$ \\ 
\textbf{PG}$(\mathrm{U}_{\text{he}}^\star)$ & $\mathbf{71.3} \,\pm\, .001$ \\ 
\bottomrule
\end{tabular}
}
\caption{\textbf{AUC.} on the binary classification task \textit{Credit default}.}
\label{tab:credit_default_auc}
\end{wraptable}

\cref{tab:uci_accuracy} shows the accuracy results, while \cref{tab:credit_default_auc} extends this to the AUC for the binary classification task. On the imbalanced \textit{Credit default} dataset, accuracy is high across methods but AUC reveals that \textbf{PG}$(\mathrm{U}_{\text{he}}^\star)$ better separates classes. The \textit{Poker} dataset remains challenging for all methods, yet \textbf{PG}$(\mathrm{U}_{\text{he}}^\star)$ still provides the best performance.

\subsection{Model-Based Reinforcement Learning}
Model-Based Reinforcement Learning (MBRL) addresses the supervised learning problem of estimating the (possibly stochastic) transition function of a MDP. Typically, we assume access to data of the form $\Dataset = \{(s^i_t, a^i_t, s^i_{t+1})\}_{i=0}^N$, consisting of trajectories of states $s$ and actions $a$ collected by an unknown policy. The goal is to approximate the next-state distribution $S_{t+1} \mid S_t, A_t \sim q$. In practice, the dynamics model is often a Gaussian probabilistic model trained via log-likelihood~\citep{Chua2018, Janner2019}, which makes it directly applicable to our experimental setup. We consider three D4RL~\citep{fu2021d4rl} \textit{HalfCheetah} tasks, each from a different data-collecting policy: \emph{simple}, \emph{medium}, and \emph{expert}, accessible through the \textit{Minari} project~\citep{minari}. All models train for $400$ epochs with Adam optimizer (learning rate = $10^{-3}$) and $\lambda = 1$.

\begin{wraptable}[25]{r}{0.5\textwidth}
    \centering
    \begin{tabular}[t]{lcc}
        \toprule
        \multirow{2}{*}{\textbf{Task}} & \multicolumn{2}{c}{\textbf{Metrics}} \\
        \cmidrule(lr){2-3}
         & \textbf{MSE}\tiny{$/ 10^{-2}\downarrow$} & \textbf{NLL}\tiny{$/ 10^{-2}\downarrow$} \\
        \midrule
        \rowcolor{red!15}
        \multicolumn{3}{l}{\textbf{NLL}} \\
        simple & 425 $\pm$ 3 & \textbf{47 $\pm$ 1} \\
        medium & 459 $\pm$ 3 & \textbf{73 $\pm$ 1} \\
        expert & 539 $\pm$ 3 & \textbf{49 $\pm$ 1} \\
        \rowcolor{blueseaborn!15}
        \multicolumn{3}{l}{\textbf{PG$(\identity_n)$}} \\
        simple & 199 $\pm$ 1 & 528 $\pm$ 18 \\
        medium & 241 $\pm$ 4 & 796 $\pm$ 70 \\
        expert & \textbf{174 $\pm$ 3} & 420 $\pm$ 24 \\
        \rowcolor{orangeseaborn!15}
        \multicolumn{3}{l}{\textbf{PG}$(\mathrm{U}_{\text{he}}^\star)$} \\
        simple & 230 $\pm$ 1 & 267 $\pm$ 1 \\
        medium & 274 $\pm$ 1 & 286 $\pm$ 1 \\
        expert & \underline{198 $\pm$ 1} & 290 $\pm$ 1 \\
        \rowcolor{greenseaborn!15}
        \multicolumn{3}{l}{\textbf{PG}$(\mathrm{U}_{\text{im}}^\star)$} \\
        simple & \textbf{190 $\pm$ 1} & \underline{208 $\pm$ 1} \\
        medium & \textbf{231 $\pm$ 1} & \underline{232 $\pm$ 2} \\
        expert & \textbf{176 $\pm$ 1} & \underline{208 $\pm$ 2} \\
        \bottomrule
    \end{tabular}
    \caption{\textbf{MBRL experiment.} The \ac{PG} loss with optimal reward comes \underline{second} to the NLL baseline in terms of NLL, and ranks \textbf{first} in terms of MSE.}
    \label{table:mbrl}
\end{wraptable}

\cref{table:mbrl} presents MSE and NLL results across the different losses under evaluation. As expected, NLL-optimized models achieve the strongest performance on NLL. However, and consistent with the synthetic data experiments in \cref{subsec:empirical}, we observe that the \ac{PG} loss with the optimal reward heuristic \textbf{PG}$(\mathrm{U}_{\text{he}}^\star)$ delivers significant NLL improvements compared to \ac{PG} with the negative squared-distance reward \textbf{PG$(\identity_n)$}. Moreover, the optimal reward obtained from the implicit differentiation solver \textbf{PG}$(\mathrm{U}_{\text{im}}^\star)$ achieves the second-best NLL performance while simultaneously achieving optimal MSE, an important property in the context of MBRL, particularly when using deterministic planners. 

These findings support our intuition that \ac{PG} methods with an optimal reward can enhance NLL (in this case also MSE), as guaranteed by our bilevel optimization framework. It is worth emphasizing that, in the context of MBRL, the metric of ultimate interest is the policy performance derived from these models, typically quantified by the return (\eg, the cumulative reward up to the task horizon). We defer exploration of this direction to future work, as the focus in the present paper is on the \ac{MLE} task.

\section{Conclusion}
\label{sec:conclusion}

In this paper, we investigated how to learn reward functions that, when used within a policy gradient algorithm, produce models that are optimal in the sense of maximum likelihood with respect to observed data. To address this question, we introduced a bilevel optimization framework and derived closed-form solutions under specific assumptions on the reward model and the data-generating distribution. Finally, we validated our approach against practical applications, showing that our framework facilitates a more effective use of the advantages of \ac{PG} methods through an optimal choice of the reward function.

\paragraph{Limitations \& Future Work.} The reward parametrization used in our work is somewhat restrictive, potentially limiting its flexibility across a wider range of tasks. While we have validated the framework in both synthetic and practical environments, further large-scale experiments are necessary to more thoroughly assess its generalizability to complex applications. Additionally, our experiments have predominantly focused on tabular data, we therefore aim to extend our approach to domains where Maximum Likelihood Estimation is known to encounter challenges, such as compounding errors, exposure bias, and limited exploration. These include areas like LLM fine-tuning, structured prediction tasks (\eg, machine translation), and time series forecasting. We intend to actively explore these directions in future research.


\section*{Acknowledgements}
The authors extend their gratitude to Ambroise Odonnat, Hamza Cherkaoui, and Michael Arbel for insightful discussions on the initial drafts of this project. This work was made possible thanks to open-source software, including Python~\citep{van1995python}, PyTorch~\citep{pytorch}, and TorchOpt~\citep{JMLR:TorchOpt}.

\section*{Reproducibility Statement}
In order to ensure reproducibility we release the code at \href{https://github.com/abenechehab/nll_to_po}{https://github.com/abenechehab/nll\_to\_po}. Implementation details and relevant hyperparameters are provided in each experiment section of the main text.

\bibliography{main}

\begin{thebibliography}{112}
\providecommand{\natexlab}[1]{#1}
\providecommand{\url}[1]{\texttt{#1}}
\expandafter\ifx\csname urlstyle\endcsname\relax
  \providecommand{\doi}[1]{doi: #1}\else
  \providecommand{\doi}{doi: \begingroup \urlstyle{rm}\Url}\fi

\bibitem[Abbeel \& Ng(2004)Abbeel and Ng]{abbeel2004}
Pieter Abbeel and Andrew~Y. Ng.
\newblock Apprenticeship learning via inverse reinforcement learning.
\newblock In \emph{Proceedings of the Twenty-First International Conference on Machine Learning}, ICML '04, pp.\ ~1, New York, NY, USA, 2004. Association for Computing Machinery.
\newblock ISBN 1581138385.
\newblock \doi{10.1145/1015330.1015430}.
\newblock URL \url{https://doi.org/10.1145/1015330.1015430}.

\bibitem[Akaike(1973)]{Akaike73}
H~Akaike.
\newblock {Information Theory and an Extension of the Maximum Likelihood Principle}.
\newblock In \emph{2nd International Symposium on Information Theory, 1973}, pp.\  268--281. Publishing House of the Hungarian Academy of Sciences, 1973.

\bibitem[Arbel \& Mairal(2022)Arbel and Mairal]{arbel2022amortized}
Michael Arbel and Julien Mairal.
\newblock Amortized implicit differentiation for stochastic bilevel optimization.
\newblock In \emph{International Conference on Learning Representations}, 2022.
\newblock URL \url{https://openreview.net/forum?id=3PN4iyXBeF}.

\bibitem[Arjovsky et~al.(2017)Arjovsky, Chintala, and Bottou]{arjovsky2017}
Martin Arjovsky, Soumith Chintala, and Léon Bottou.
\newblock Wasserstein gan, 2017.
\newblock URL \url{https://arxiv.org/abs/1701.07875}.

\bibitem[Bahdanau et~al.(2017)Bahdanau, Brakel, Xu, Goyal, Lowe, Pineau, Courville, and Bengio]{bahdanau2017}
Dzmitry Bahdanau, Philemon Brakel, Kelvin Xu, Anirudh Goyal, Ryan Lowe, Joelle Pineau, Aaron Courville, and Yoshua Bengio.
\newblock An actor-critic algorithm for sequence prediction, 2017.
\newblock URL \url{https://arxiv.org/abs/1607.07086}.

\bibitem[Bai et~al.(2025)Bai, Chen, Liu, Wang, Ge, Song, Dang, Wang, Wang, Tang, Zhong, Zhu, Yang, Li, Wan, Wang, Ding, Fu, Xu, Ye, Zhang, Xie, Cheng, Zhang, Yang, Xu, and Lin]{bai2025qwen25vl}
Shuai Bai, Keqin Chen, Xuejing Liu, Jialin Wang, Wenbin Ge, Sibo Song, Kai Dang, Peng Wang, Shijie Wang, Jun Tang, Humen Zhong, Yuanzhi Zhu, Mingkun Yang, Zhaohai Li, Jianqiang Wan, Pengfei Wang, Wei Ding, Zheren Fu, Yiheng Xu, Jiabo Ye, Xi~Zhang, Tianbao Xie, Zesen Cheng, Hang Zhang, Zhibo Yang, Haiyang Xu, and Junyang Lin.
\newblock Qwen2.5-vl technical report, 2025.
\newblock URL \url{https://arxiv.org/abs/2502.13923}.

\bibitem[Bao et~al.(2020)Bao, Li, Xu, Su, Zhu, and Zhang]{bao2020}
Fan Bao, Chongxuan Li, Kun Xu, Hang Su, Jun Zhu, and Bo~Zhang.
\newblock Bi-level score matching for learning energy-based latent variable models.
\newblock In \emph{Proceedings of the 34th International Conference on Neural Information Processing Systems}, NIPS '20, Red Hook, NY, USA, 2020. Curran Associates Inc.
\newblock ISBN 9781713829546.

\bibitem[Bellman(1958)]{bellman1958}
Richard Bellman.
\newblock Dynamic programming and stochastic control processes.
\newblock \emph{Information and Control}, 1\penalty0 (3):\penalty0 228--239, 1958.
\newblock ISSN 0019-9958.
\newblock \doi{https://doi.org/10.1016/S0019-9958(58)80003-0}.
\newblock URL \url{https://www.sciencedirect.com/science/article/pii/S0019995858800030}.

\bibitem[Benechehab et~al.(2024)Benechehab, Thomas, Paolo, Filippone, and Kégl]{benechehab2024}
Abdelhakim Benechehab, Albert Thomas, Giuseppe Paolo, Maurizio Filippone, and Balázs Kégl.
\newblock A multi-step loss function for robust learning of the dynamics in model-based reinforcement learning, 2024.
\newblock URL \url{https://arxiv.org/abs/2402.03146}.

\bibitem[Benechehab et~al.(2025)Benechehab, Hili, Odonnat, Zekri, Thomas, Paolo, Filippone, Redko, and Kégl]{benechehab2025zeroshot}
Abdelhakim Benechehab, Youssef Attia~El Hili, Ambroise Odonnat, Oussama Zekri, Albert Thomas, Giuseppe Paolo, Maurizio Filippone, Ievgen Redko, and Balázs Kégl.
\newblock Zero-shot model-based reinforcement learning using large language models.
\newblock In \emph{The Thirteenth International Conference on Learning Representations}, 2025.
\newblock URL \url{https://openreview.net/forum?id=uZFXpPrwSh}.

\bibitem[Bengio et~al.(2015)Bengio, Vinyals, Jaitly, and Shazeer]{bengio2015sch}
Samy Bengio, Oriol Vinyals, Navdeep Jaitly, and Noam Shazeer.
\newblock Scheduled sampling for sequence prediction with recurrent neural networks, 2015.
\newblock URL \url{https://arxiv.org/abs/1506.03099}.

\bibitem[Bengio(2000)]{bengio2000}
Yoshua Bengio.
\newblock Gradient-based optimization of hyperparameters.
\newblock \emph{Neural Computation}, 12\penalty0 (8):\penalty0 1889--1900, 2000.
\newblock \doi{10.1162/089976600300015187}.

\bibitem[Bennett et~al.(2006)Bennett, Hu, Ji, Kunapuli, and Pang]{bennett2006}
K.P. Bennett, Jing Hu, Xiaoyun Ji, G.~Kunapuli, and Jong-Shi Pang.
\newblock Model selection via bilevel optimization.
\newblock In \emph{The 2006 IEEE International Joint Conference on Neural Network Proceedings}, pp.\  1922--1929, 2006.
\newblock \doi{10.1109/IJCNN.2006.246935}.

\bibitem[Black et~al.(2024)Black, Janner, Du, Kostrikov, and Levine]{black2024training}
Kevin Black, Michael Janner, Yilun Du, Ilya Kostrikov, and Sergey Levine.
\newblock Training diffusion models with reinforcement learning.
\newblock In \emph{The Twelfth International Conference on Learning Representations}, 2024.
\newblock URL \url{https://openreview.net/forum?id=YCWjhGrJFD}.

\bibitem[Bradley \& Terry(1952)Bradley and Terry]{bradley_terry1952}
Ralph~Allan Bradley and Milton~E. Terry.
\newblock Rank analysis of incomplete block designs: I. the method of paired comparisons.
\newblock \emph{Biometrika}, 39\penalty0 (3/4):\penalty0 324--345, 1952.
\newblock ISSN 00063444, 14643510.
\newblock URL \url{http://www.jstor.org/stable/2334029}.

\bibitem[Brown et~al.(2020)Brown, Mann, Ryder, Subbiah, Kaplan, Dhariwal, Neelakantan, Shyam, Sastry, Askell, Agarwal, Herbert-Voss, Krueger, Henighan, Child, Ramesh, Ziegler, Wu, Winter, Hesse, Chen, Sigler, Litwin, Gray, Chess, Clark, Berner, McCandlish, Radford, Sutskever, and Amodei]{brown2020languagemodelsfewshotlearners}
Tom~B. Brown, Benjamin Mann, Nick Ryder, Melanie Subbiah, Jared Kaplan, Prafulla Dhariwal, Arvind Neelakantan, Pranav Shyam, Girish Sastry, Amanda Askell, Sandhini Agarwal, Ariel Herbert-Voss, Gretchen Krueger, Tom Henighan, Rewon Child, Aditya Ramesh, Daniel~M. Ziegler, Jeffrey Wu, Clemens Winter, Christopher Hesse, Mark Chen, Eric Sigler, Mateusz Litwin, Scott Gray, Benjamin Chess, Jack Clark, Christopher Berner, Sam McCandlish, Alec Radford, Ilya Sutskever, and Dario Amodei.
\newblock Language models are few-shot learners, 2020.
\newblock URL \url{https://arxiv.org/abs/2005.14165}.

\bibitem[Chakraborty et~al.(2024)Chakraborty, Bedi, Koppel, Wang, Manocha, Wang, and Huang]{chakraborty2024}
Souradip Chakraborty, Amrit Bedi, Alec Koppel, Huazheng Wang, Dinesh Manocha, Mengdi Wang, and Furong Huang.
\newblock {PARL}: A unified framework for policy alignment in reinforcement learning from human feedback.
\newblock In \emph{The Twelfth International Conference on Learning Representations}, 2024.
\newblock URL \url{https://openreview.net/forum?id=ByR3NdDSZB}.

\bibitem[Chen et~al.(2021)Chen, Sun, and Yin]{chen2021closing}
Tianyi Chen, Yuejiao Sun, and Wotao Yin.
\newblock Closing the gap: Tighter analysis of alternating stochastic gradient methods for bilevel problems.
\newblock In M.~Ranzato, A.~Beygelzimer, Y.~Dauphin, P.S. Liang, and J.~Wortman Vaughan (eds.), \emph{Advances in Neural Information Processing Systems}, volume~34, pp.\  25294--25307. Curran Associates, Inc., 2021.
\newblock URL \url{https://proceedings.neurips.cc/paper_files/paper/2021/file/d4dd111a4fd973394238aca5c05bebe3-Paper.pdf}.

\bibitem[Christiano et~al.(2017)Christiano, Leike, Brown, Martic, Legg, and Amodei]{christiano2017}
Paul~F Christiano, Jan Leike, Tom Brown, Miljan Martic, Shane Legg, and Dario Amodei.
\newblock Deep reinforcement learning from human preferences.
\newblock In I.~Guyon, U.~Von Luxburg, S.~Bengio, H.~Wallach, R.~Fergus, S.~Vishwanathan, and R.~Garnett (eds.), \emph{Advances in Neural Information Processing Systems}, volume~30. Curran Associates, Inc., 2017.
\newblock URL \url{https://proceedings.neurips.cc/paper_files/paper/2017/file/d5e2c0adad503c91f91df240d0cd4e49-Paper.pdf}.

\bibitem[Chua et~al.(2018)Chua, Calandra, McAllister, and Levine]{Chua2018}
Kurtland Chua, Roberto Calandra, Rowan McAllister, and Sergey Levine.
\newblock Deep reinforcement learning in a handful of trials using probabilistic dynamics models.
\newblock In \emph{Advances in Neural Information Processing Systems 31}, pp.\  4754--4765. Curran Associates, Inc., 2018.

\bibitem[Dagréou et~al.(2024)Dagréou, Ablin, Vaiter, and Moreau]{dagreou2024}
Mathieu Dagréou, Pierre Ablin, Samuel Vaiter, and Thomas Moreau.
\newblock A framework for bilevel optimization that enables stochastic and global variance reduction algorithms, 2024.
\newblock URL \url{https://arxiv.org/abs/2201.13409}.

\bibitem[Dai et~al.(2024)Dai, Pan, Sun, Ji, Xu, Liu, Wang, and Yang]{dai2024safe}
Josef Dai, Xuehai Pan, Ruiyang Sun, Jiaming Ji, Xinbo Xu, Mickel Liu, Yizhou Wang, and Yaodong Yang.
\newblock Safe {RLHF}: Safe reinforcement learning from human feedback.
\newblock In \emph{The Twelfth International Conference on Learning Representations}, 2024.
\newblock URL \url{https://openreview.net/forum?id=TyFrPOKYXw}.

\bibitem[DeepSeek-AI et~al.(2025)DeepSeek-AI, Guo, Yang, Zhang, Song, Zhang, Xu, Zhu, Ma, Wang, Bi, Zhang, Yu, Wu, Wu, Gou, Shao, Li, Gao, Liu, Xue, Wang, Wu, Feng, Lu, Zhao, Deng, Zhang, Ruan, Dai, Chen, Ji, Li, Lin, Dai, Luo, Hao, Chen, Li, Zhang, Bao, Xu, Wang, Ding, Xin, Gao, Qu, Li, Guo, Li, Wang, Chen, Yuan, Qiu, Li, Cai, Ni, Liang, Chen, Dong, Hu, Gao, Guan, Huang, Yu, Wang, Zhang, Zhao, Wang, Zhang, Xu, Xia, Zhang, Zhang, Tang, Li, Wang, Li, Tian, Huang, Zhang, Wang, Chen, Du, Ge, Zhang, Pan, Wang, Chen, Jin, Chen, Lu, Zhou, Chen, Ye, Wang, Yu, Zhou, Pan, Li, Zhou, Wu, Ye, Yun, Pei, Sun, Wang, Zeng, Zhao, Liu, Liang, Gao, Yu, Zhang, Xiao, An, Liu, Wang, Chen, Nie, Cheng, Liu, Xie, Liu, Yang, Li, Su, Lin, Li, Jin, Shen, Chen, Sun, Wang, Song, Zhou, Wang, Shan, Li, Wang, Wei, Zhang, Xu, Li, Zhao, Sun, Wang, Yu, Zhang, Shi, Xiong, He, Piao, Wang, Tan, Ma, Liu, Guo, Ou, Wang, Gong, Zou, He, Xiong, Luo, You, Liu, Zhou, Zhu, Xu, Huang, Li, Zheng, Zhu, Ma, Tang, Zha, Yan, Ren, Ren, Sha, Fu, Xu, Xie, Zhang,
  Hao, Ma, Yan, Wu, Gu, Zhu, Liu, Li, Xie, Song, Pan, Huang, Xu, Zhang, and Zhang]{deepseekr1}
DeepSeek-AI, Daya Guo, Dejian Yang, Haowei Zhang, Junxiao Song, Ruoyu Zhang, Runxin Xu, Qihao Zhu, Shirong Ma, Peiyi Wang, Xiao Bi, Xiaokang Zhang, Xingkai Yu, Yu~Wu, Z.~F. Wu, Zhibin Gou, Zhihong Shao, Zhuoshu Li, Ziyi Gao, Aixin Liu, Bing Xue, Bingxuan Wang, Bochao Wu, Bei Feng, Chengda Lu, Chenggang Zhao, Chengqi Deng, Chenyu Zhang, Chong Ruan, Damai Dai, Deli Chen, Dongjie Ji, Erhang Li, Fangyun Lin, Fucong Dai, Fuli Luo, Guangbo Hao, Guanting Chen, Guowei Li, H.~Zhang, Han Bao, Hanwei Xu, Haocheng Wang, Honghui Ding, Huajian Xin, Huazuo Gao, Hui Qu, Hui Li, Jianzhong Guo, Jiashi Li, Jiawei Wang, Jingchang Chen, Jingyang Yuan, Junjie Qiu, Junlong Li, J.~L. Cai, Jiaqi Ni, Jian Liang, Jin Chen, Kai Dong, Kai Hu, Kaige Gao, Kang Guan, Kexin Huang, Kuai Yu, Lean Wang, Lecong Zhang, Liang Zhao, Litong Wang, Liyue Zhang, Lei Xu, Leyi Xia, Mingchuan Zhang, Minghua Zhang, Minghui Tang, Meng Li, Miaojun Wang, Mingming Li, Ning Tian, Panpan Huang, Peng Zhang, Qiancheng Wang, Qinyu Chen, Qiushi Du, Ruiqi Ge, Ruisong
  Zhang, Ruizhe Pan, Runji Wang, R.~J. Chen, R.~L. Jin, Ruyi Chen, Shanghao Lu, Shangyan Zhou, Shanhuang Chen, Shengfeng Ye, Shiyu Wang, Shuiping Yu, Shunfeng Zhou, Shuting Pan, S.~S. Li, Shuang Zhou, Shaoqing Wu, Shengfeng Ye, Tao Yun, Tian Pei, Tianyu Sun, T.~Wang, Wangding Zeng, Wanjia Zhao, Wen Liu, Wenfeng Liang, Wenjun Gao, Wenqin Yu, Wentao Zhang, W.~L. Xiao, Wei An, Xiaodong Liu, Xiaohan Wang, Xiaokang Chen, Xiaotao Nie, Xin Cheng, Xin Liu, Xin Xie, Xingchao Liu, Xinyu Yang, Xinyuan Li, Xuecheng Su, Xuheng Lin, X.~Q. Li, Xiangyue Jin, Xiaojin Shen, Xiaosha Chen, Xiaowen Sun, Xiaoxiang Wang, Xinnan Song, Xinyi Zhou, Xianzu Wang, Xinxia Shan, Y.~K. Li, Y.~Q. Wang, Y.~X. Wei, Yang Zhang, Yanhong Xu, Yao Li, Yao Zhao, Yaofeng Sun, Yaohui Wang, Yi~Yu, Yichao Zhang, Yifan Shi, Yiliang Xiong, Ying He, Yishi Piao, Yisong Wang, Yixuan Tan, Yiyang Ma, Yiyuan Liu, Yongqiang Guo, Yuan Ou, Yuduan Wang, Yue Gong, Yuheng Zou, Yujia He, Yunfan Xiong, Yuxiang Luo, Yuxiang You, Yuxuan Liu, Yuyang Zhou, Y.~X. Zhu,
  Yanhong Xu, Yanping Huang, Yaohui Li, Yi~Zheng, Yuchen Zhu, Yunxian Ma, Ying Tang, Yukun Zha, Yuting Yan, Z.~Z. Ren, Zehui Ren, Zhangli Sha, Zhe Fu, Zhean Xu, Zhenda Xie, Zhengyan Zhang, Zhewen Hao, Zhicheng Ma, Zhigang Yan, Zhiyu Wu, Zihui Gu, Zijia Zhu, Zijun Liu, Zilin Li, Ziwei Xie, Ziyang Song, Zizheng Pan, Zhen Huang, Zhipeng Xu, Zhongyu Zhang, and Zhen Zhang.
\newblock Deepseek-r1: Incentivizing reasoning capability in llms via reinforcement learning, 2025.
\newblock URL \url{https://arxiv.org/abs/2501.12948}.

\bibitem[Ding \& Soricut(2017)Ding and Soricut]{Ding2017ColdStartRL}
Nan Ding and Radu Soricut.
\newblock Cold-start reinforcement learning with softmax policy gradient.
\newblock In \emph{Neural Information Processing Systems}, 2017.
\newblock URL \url{https://api.semanticscholar.org/CorpusID:4206469}.

\bibitem[Domke(2012)]{domke2012}
Justin Domke.
\newblock Generic methods for optimization-based modeling.
\newblock In Neil~D. Lawrence and Mark Girolami (eds.), \emph{Proceedings of the Fifteenth International Conference on Artificial Intelligence and Statistics}, volume~22 of \emph{Proceedings of Machine Learning Research}, pp.\  318--326, La Palma, Canary Islands, 21--23 Apr 2012. PMLR.
\newblock URL \url{https://proceedings.mlr.press/v22/domke12.html}.

\bibitem[Fan et~al.(2022)Fan, Wang, Jiang, Mandlekar, Yang, Zhu, Tang, Huang, Zhu, and Anandkumar]{fan2022minedojo}
Linxi Fan, Guanzhi Wang, Yunfan Jiang, Ajay Mandlekar, Yuncong Yang, Haoyi Zhu, Andrew Tang, De-An Huang, Yuke Zhu, and Anima Anandkumar.
\newblock Minedojo: Building open-ended embodied agents with internet-scale knowledge.
\newblock In \emph{Thirty-sixth Conference on Neural Information Processing Systems Datasets and Benchmarks Track}, 2022.
\newblock URL \url{https://openreview.net/forum?id=rc8o_j8I8PX}.

\bibitem[Feng et~al.(2021)Feng, Slumbers, Wan, Liu, McAleer, Wen, Wang, and Yang]{feng2021}
Xidong Feng, Oliver Slumbers, Ziyu Wan, Bo~Liu, Stephen McAleer, Ying Wen, Jun Wang, and Yaodong Yang.
\newblock Neural auto-curricula, 2021.
\newblock URL \url{https://arxiv.org/abs/2106.02745}.

\bibitem[Finn et~al.(2016{\natexlab{a}})Finn, Christiano, Abbeel, and Levine]{finn2016connection}
Chelsea Finn, Paul Christiano, Pieter Abbeel, and Sergey Levine.
\newblock A connection between generative adversarial networks, inverse reinforcement learning, and energy-based models, 2016{\natexlab{a}}.
\newblock URL \url{https://arxiv.org/abs/1611.03852}.

\bibitem[Finn et~al.(2016{\natexlab{b}})Finn, Levine, and Abbeel]{finn2016guided}
Chelsea Finn, Sergey Levine, and Pieter Abbeel.
\newblock Guided cost learning: Deep inverse optimal control via policy optimization.
\newblock In Maria~Florina Balcan and Kilian~Q. Weinberger (eds.), \emph{Proceedings of The 33rd International Conference on Machine Learning}, volume~48 of \emph{Proceedings of Machine Learning Research}, pp.\  49--58, New York, New York, USA, 20--22 Jun 2016{\natexlab{b}}. PMLR.
\newblock URL \url{https://proceedings.mlr.press/v48/finn16.html}.

\bibitem[Franceschi et~al.(2017)Franceschi, Donini, Frasconi, and Pontil]{franceschi2017}
Luca Franceschi, Michele Donini, Paolo Frasconi, and Massimiliano Pontil.
\newblock Forward and reverse gradient-based hyperparameter optimization.
\newblock In Doina Precup and Yee~Whye Teh (eds.), \emph{Proceedings of the 34th International Conference on Machine Learning}, volume~70 of \emph{Proceedings of Machine Learning Research}, pp.\  1165--1173. PMLR, 06--11 Aug 2017.
\newblock URL \url{https://proceedings.mlr.press/v70/franceschi17a.html}.

\bibitem[Franceschi et~al.(2018)Franceschi, Frasconi, Salzo, Grazzi, and Pontil]{franceschi2018}
Luca Franceschi, Paolo Frasconi, Saverio Salzo, Riccardo Grazzi, and Massimiliano Pontil.
\newblock Bilevel programming for hyperparameter optimization and meta-learning.
\newblock In Jennifer Dy and Andreas Krause (eds.), \emph{Proceedings of the 35th International Conference on Machine Learning}, volume~80 of \emph{Proceedings of Machine Learning Research}, pp.\  1568--1577. PMLR, 10--15 Jul 2018.
\newblock URL \url{https://proceedings.mlr.press/v80/franceschi18a.html}.

\bibitem[Fu et~al.(2025)Fu, Lin, Wang, Zhang, Shen, Liu, Cao, Long, Gao, Li, Ma, Zheng, Ji, Sun, Shan, and He]{fu2025vita}
Chaoyou Fu, Haojia Lin, Xiong Wang, Yi-Fan Zhang, Yunhang Shen, Xiaoyu Liu, Haoyu Cao, Zuwei Long, Heting Gao, Ke~Li, Long Ma, Xiawu Zheng, Rongrong Ji, Xing Sun, Caifeng Shan, and Ran He.
\newblock Vita-1.5: Towards gpt-4o level real-time vision and speech interaction, 2025.
\newblock URL \url{https://arxiv.org/abs/2501.01957}.

\bibitem[Fu et~al.(2021)Fu, Kumar, Nachum, Tucker, and Levine]{fu2021d4rl}
Justin Fu, Aviral Kumar, Ofir Nachum, George Tucker, and Sergey Levine.
\newblock D4rl: Datasets for deep data-driven reinforcement learning, 2021.
\newblock URL \url{https://arxiv.org/abs/2004.07219}.

\bibitem[Gaur et~al.(2025)Gaur, Singh, Bedi, Pasupathu, and Aggarwal]{gaur2025}
Mudit Gaur, Utsav Singh, Amrit~Singh Bedi, Raghu Pasupathu, and Vaneet Aggarwal.
\newblock On the sample complexity bounds in bilevel reinforcement learning, 2025.
\newblock URL \url{https://arxiv.org/abs/2503.17644}.

\bibitem[Goodfellow et~al.(2014)Goodfellow, Pouget-Abadie, Mirza, Xu, Warde-Farley, Ozair, Courville, and Bengio]{goodfellow2014}
Ian~J. Goodfellow, Jean Pouget-Abadie, Mehdi Mirza, Bing Xu, David Warde-Farley, Sherjil Ozair, Aaron Courville, and Yoshua Bengio.
\newblock Generative adversarial networks, 2014.
\newblock URL \url{https://arxiv.org/abs/1406.2661}.

\bibitem[Greensmith et~al.(2001)Greensmith, Bartlett, and Baxter]{greensmith2001}
Evan Greensmith, Peter Bartlett, and Jonathan Baxter.
\newblock Variance reduction techniques for gradient estimates in reinforcement learning.
\newblock In T.~Dietterich, S.~Becker, and Z.~Ghahramani (eds.), \emph{Advances in Neural Information Processing Systems}, volume~14. MIT Press, 2001.
\newblock URL \url{https://proceedings.neurips.cc/paper_files/paper/2001/file/584b98aac2dddf59ee2cf19ca4ccb75e-Paper.pdf}.

\bibitem[Haarnoja et~al.(2017)Haarnoja, Tang, Abbeel, and Levine]{haarnoja2017reinforcement}
Tuomas Haarnoja, Haoran Tang, Pieter Abbeel, and Sergey Levine.
\newblock Reinforcement learning with deep energy-based policies.
\newblock 2017.

\bibitem[Haarnoja et~al.(2018)Haarnoja, Zhou, Abbeel, and Levine]{haarnoja2018sac}
Tuomas Haarnoja, Aurick Zhou, Pieter Abbeel, and Sergey Levine.
\newblock Soft actor-critic: Off-policy maximum entropy deep reinforcement learning with a stochastic actor, 2018.
\newblock URL \url{https://arxiv.org/abs/1801.01290}.

\bibitem[Hafner et~al.(2021)Hafner, Lillicrap, Norouzi, and Ba]{Hafner2021}
Danijar Hafner, Timothy~P Lillicrap, Mohammad Norouzi, and Jimmy Ba.
\newblock Mastering atari with discrete world models.
\newblock In \emph{International Conference on Learning Representations}, 2021.
\newblock URL \url{https://openreview.net/forum?id=0oabwyZbOu}.

\bibitem[Hartikainen et~al.(2020)Hartikainen, Geng, Haarnoja, and Levine]{Hartikainen2020Dynamical}
Kristian Hartikainen, Xinyang Geng, Tuomas Haarnoja, and Sergey Levine.
\newblock Dynamical distance learning for semi-supervised and unsupervised skill discovery.
\newblock In \emph{International Conference on Learning Representations}, 2020.
\newblock URL \url{https://openreview.net/forum?id=H1lmhaVtvr}.

\bibitem[Higgins et~al.(2017)Higgins, Matthey, Pal, Burgess, Glorot, Botvinick, Mohamed, and Lerchner]{higgins2017betavae}
Irina Higgins, Loic Matthey, Arka Pal, Christopher Burgess, Xavier Glorot, Matthew Botvinick, Shakir Mohamed, and Alexander Lerchner.
\newblock beta-{VAE}: Learning basic visual concepts with a constrained variational framework.
\newblock In \emph{International Conference on Learning Representations}, 2017.
\newblock URL \url{https://openreview.net/forum?id=Sy2fzU9gl}.

\bibitem[Hong et~al.(2022)Hong, Wai, Wang, and Yang]{hong2022}
Mingyi Hong, Hoi-To Wai, Zhaoran Wang, and Zhuoran Yang.
\newblock A two-timescale framework for bilevel optimization: Complexity analysis and application to actor-critic, 2022.
\newblock URL \url{https://arxiv.org/abs/2007.05170}.

\bibitem[Intelligence et~al.(2025)Intelligence, Black, Brown, Darpinian, Dhabalia, Driess, Esmail, Equi, Finn, Fusai, Galliker, Ghosh, Groom, Hausman, Ichter, Jakubczak, Jones, Ke, LeBlanc, Levine, Li-Bell, Mothukuri, Nair, Pertsch, Ren, Shi, Smith, Springenberg, Stachowicz, Tanner, Vuong, Walke, Walling, Wang, Yu, and Zhilinsky]{intelligence2025}
Physical Intelligence, Kevin Black, Noah Brown, James Darpinian, Karan Dhabalia, Danny Driess, Adnan Esmail, Michael Equi, Chelsea Finn, Niccolo Fusai, Manuel~Y. Galliker, Dibya Ghosh, Lachy Groom, Karol Hausman, Brian Ichter, Szymon Jakubczak, Tim Jones, Liyiming Ke, Devin LeBlanc, Sergey Levine, Adrian Li-Bell, Mohith Mothukuri, Suraj Nair, Karl Pertsch, Allen~Z. Ren, Lucy~Xiaoyang Shi, Laura Smith, Jost~Tobias Springenberg, Kyle Stachowicz, James Tanner, Quan Vuong, Homer Walke, Anna Walling, Haohuan Wang, Lili Yu, and Ury Zhilinsky.
\newblock $\pi_{0.5}$: a vision-language-action model with open-world generalization, 2025.
\newblock URL \url{https://arxiv.org/abs/2504.16054}.

\bibitem[Janner et~al.(2019)Janner, Fu, Zhang, and Levine]{Janner2019}
Michael Janner, Justin Fu, Marvin Zhang, and Sergey Levine.
\newblock When to trust your model: Model-based policy optimization.
\newblock In H.~Wallach, H.~Larochelle, A.~Beygelzimer, F.~d\textquotesingle Alch\'{e}-Buc, E.~Fox, and R.~Garnett (eds.), \emph{Advances in Neural Information Processing Systems}, volume~32. Curran Associates, Inc., 2019.

\bibitem[Ji et~al.(2021)Ji, Yang, and Liang]{ji2021}
Kaiyi Ji, Junjie Yang, and Yingbin Liang.
\newblock Bilevel optimization: Convergence analysis and enhanced design, 2021.
\newblock URL \url{https://arxiv.org/abs/2010.07962}.

\bibitem[Kan et~al.(2022)Kan, Lü, Wang, Zhang, Zhu, Huang, Guo, and Snoussi]{kan2022}
Ge~Kan, Jinhu Lü, Tian Wang, Baochang Zhang, Aichun Zhu, Lei Huang, Guodong Guo, and Hichem Snoussi.
\newblock Bi-level doubly variational learning for energy-based latent variable models, 2022.
\newblock URL \url{https://arxiv.org/abs/2203.14702}.

\bibitem[K{\'e}gl et~al.(2021)K{\'e}gl, Hurtado, and Thomas]{Kegl2021}
Bal{\'a}zs K{\'e}gl, Gabriel Hurtado, and Albert Thomas.
\newblock Model-based micro-data reinforcement learning: what are the crucial model properties and which model to choose?
\newblock In \emph{International Conference on Learning Representations}, 2021.
\newblock URL \url{https://openreview.net/forum?id=p5uylG94S68}.

\bibitem[Kim et~al.(2023)Kim, Park, Shin, Lee, Abbeel, and Lee]{kim2023preference}
Changyeon Kim, Jongjin Park, Jinwoo Shin, Honglak Lee, Pieter Abbeel, and Kimin Lee.
\newblock Preference transformer: Modeling human preferences using transformers for {RL}.
\newblock In \emph{The Eleventh International Conference on Learning Representations}, 2023.
\newblock URL \url{https://openreview.net/forum?id=Peot1SFDX0}.

\bibitem[Kim et~al.(2024)Kim, Pertsch, Karamcheti, Xiao, Balakrishna, Nair, Rafailov, Foster, Lam, Sanketi, Vuong, Kollar, Burchfiel, Tedrake, Sadigh, Levine, Liang, and Finn]{kim2024}
Moo~Jin Kim, Karl Pertsch, Siddharth Karamcheti, Ted Xiao, Ashwin Balakrishna, Suraj Nair, Rafael Rafailov, Ethan Foster, Grace Lam, Pannag Sanketi, Quan Vuong, Thomas Kollar, Benjamin Burchfiel, Russ Tedrake, Dorsa Sadigh, Sergey Levine, Percy Liang, and Chelsea Finn.
\newblock Openvla: An open-source vision-language-action model, 2024.
\newblock URL \url{https://arxiv.org/abs/2406.09246}.

\bibitem[Kingma \& Ba(2017)Kingma and Ba]{kingma2017}
Diederik~P. Kingma and Jimmy Ba.
\newblock Adam: A method for stochastic optimization, 2017.
\newblock URL \url{https://arxiv.org/abs/1412.6980}.

\bibitem[Kingma \& Welling(2013)Kingma and Welling]{Kingma2013AutoEncodingVB}
Diederik~P. Kingma and Max Welling.
\newblock Auto-encoding variational bayes.
\newblock \emph{CoRR}, abs/1312.6114, 2013.
\newblock URL \url{https://api.semanticscholar.org/CorpusID:216078090}.

\bibitem[Lai et~al.(2025)Lai, Zhao, Feng, Ma, Liu, Zhao, Lin, Yi, Xie, Zhang, Liu, Meng, and Zhu]{lai2025}
Song Lai, Haohan Zhao, Rong Feng, Changyi Ma, Wenzhuo Liu, Hongbo Zhao, Xi~Lin, Dong Yi, Min Xie, Qingfu Zhang, Hongbin Liu, Gaofeng Meng, and Fei Zhu.
\newblock Reinforcement fine-tuning naturally mitigates forgetting in continual post-training, 2025.
\newblock URL \url{https://arxiv.org/abs/2507.05386}.

\bibitem[Linnainmaa(1976)]{linnainmaa1976}
Seppo Linnainmaa.
\newblock Taylor expansion of the accumulated rounding error.
\newblock \emph{BIT}, 16\penalty0 (2):\penalty0 146–160, June 1976.
\newblock ISSN 0006-3835.
\newblock \doi{10.1007/BF01931367}.
\newblock URL \url{https://doi.org/10.1007/BF01931367}.

\bibitem[Liu et~al.(2022)Liu, Zhu, and Zhang]{liu2022goal}
Minghuan Liu, Menghui Zhu, and Weinan Zhang.
\newblock Goal-conditioned reinforcement learning: Problems and solutions, 2022.
\newblock URL \url{https://arxiv.org/abs/2201.08299}.

\bibitem[Lu et~al.(2024)Lu, Lu, Lange, Foerster, Clune, and Ha]{lu2024aiscientist}
Chris Lu, Cong Lu, Robert~Tjarko Lange, Jakob Foerster, Jeff Clune, and David Ha.
\newblock The {AI} {S}cientist: Towards fully automated open-ended scientific discovery.
\newblock \emph{arXiv preprint arXiv:2408.06292}, 2024.

\bibitem[Luong et~al.(2024)Luong, Zhang, Jie, Sun, Jin, and Li]{luong2024}
Trung~Quoc Luong, Xinbo Zhang, Zhanming Jie, Peng Sun, Xiaoran Jin, and Hang Li.
\newblock Reft: Reasoning with reinforced fine-tuning, 2024.
\newblock URL \url{https://arxiv.org/abs/2401.08967}.

\bibitem[Mackay et~al.(2019)Mackay, Vicol, Lorraine, Duvenaud, and Grosse]{mackay2018}
Matthew Mackay, Paul Vicol, Jonathan Lorraine, David Duvenaud, and Roger Grosse.
\newblock Self-tuning networks: Bilevel optimization of hyperparameters using structured best-response functions.
\newblock In \emph{International Conference on Learning Representations}, 2019.
\newblock URL \url{https://openreview.net/forum?id=r1eEG20qKQ}.

\bibitem[Mahalanobis(1936)]{mahalanobis1936}
P.C. Mahalanobis.
\newblock On the generalised distance in statistics.
\newblock \emph{{Proceedings of the National Institute of Sciences of India}}, 2:\penalty0 49--55, 1936.

\bibitem[Manica et~al.(2023)Manica, Born, Cadow, Christofidellis, Dave, Clarke, Teukam, Giannone, Hoffman, Buchan, Chenthamarakshan, Donovan, Hsu, Zipoli, Schilter, Kishimoto, Hamada, Padhi, Wehden, McHugh, Khrabrov, Das, Takeda, and Smith]{Manica2023}
Matteo Manica, Jannis Born, Joris Cadow, Dimitrios Christofidellis, Ashish Dave, Dean Clarke, Yves Gaetan~Nana Teukam, Giorgio Giannone, Samuel~C. Hoffman, Matthew Buchan, Vijil Chenthamarakshan, Timothy Donovan, Hsiang~Han Hsu, Federico Zipoli, Oliver Schilter, Akihiro Kishimoto, Lisa Hamada, Inkit Padhi, Karl Wehden, Lauren McHugh, Alexy Khrabrov, Payel Das, Seiji Takeda, and John~R. Smith.
\newblock Accelerating material design with the generative toolkit for scientific discovery.
\newblock \emph{npj Computational Materials}, 9\penalty0 (1), 2023.
\newblock ISSN 2057-3960.
\newblock \doi{10.1038/s41524-023-01028-1}.
\newblock URL \url{http://dx.doi.org/10.1038/s41524-023-01028-1}.

\bibitem[Mazzaglia et~al.(2024)Mazzaglia, Verbelen, Dhoedt, Courville, and Rajeswar]{mazzaglia2024}
Pietro Mazzaglia, Tim Verbelen, Bart Dhoedt, Aaron Courville, and Sai Rajeswar.
\newblock Genrl: Multimodal-foundation world models for generalization in embodied agents, 2024.
\newblock URL \url{https://arxiv.org/abs/2406.18043}.

\bibitem[Nachum et~al.(2018)Nachum, Gu, Lee, and Levine]{nachum2018}
Ofir Nachum, Shixiang Gu, Honglak Lee, and Sergey Levine.
\newblock Data-efficient hierarchical reinforcement learning, 2018.
\newblock URL \url{https://arxiv.org/abs/1805.08296}.

\bibitem[Nikishin et~al.(2021)Nikishin, Abachi, Agarwal, and Bacon]{nikishin2021}
Evgenii Nikishin, Romina Abachi, Rishabh Agarwal, and Pierre-Luc Bacon.
\newblock Control-oriented model-based reinforcement learning with implicit differentiation, 2021.
\newblock URL \url{https://arxiv.org/abs/2106.03273}.

\bibitem[Norouzi et~al.(2016)Norouzi, Bengio, Chen, Jaitly, Schuster, Wu, and Schuurmans]{norouzi2016}
Mohammad Norouzi, Samy Bengio, zhifeng Chen, Navdeep Jaitly, Mike Schuster, Yonghui Wu, and Dale Schuurmans.
\newblock Reward augmented maximum likelihood for neural structured prediction.
\newblock In D.~Lee, M.~Sugiyama, U.~Luxburg, I.~Guyon, and R.~Garnett (eds.), \emph{Advances in Neural Information Processing Systems}, volume~29. Curran Associates, Inc., 2016.
\newblock URL \url{https://proceedings.neurips.cc/paper_files/paper/2016/file/2f885d0fbe2e131bfc9d98363e55d1d4-Paper.pdf}.

\bibitem[Ouyang et~al.(2022)Ouyang, Wu, Jiang, Almeida, Wainwright, Mishkin, Zhang, Agarwal, Slama, Ray, Schulman, Hilton, Kelton, Miller, Simens, Askell, Welinder, Christiano, Leike, and Lowe]{ouyang2022}
Long Ouyang, Jeff Wu, Xu~Jiang, Diogo Almeida, Carroll~L. Wainwright, Pamela Mishkin, Chong Zhang, Sandhini Agarwal, Katarina Slama, Alex Ray, John Schulman, Jacob Hilton, Fraser Kelton, Luke Miller, Maddie Simens, Amanda Askell, Peter Welinder, Paul Christiano, Jan Leike, and Ryan Lowe.
\newblock Training language models to follow instructions with human feedback, 2022.
\newblock URL \url{https://arxiv.org/abs/2203.02155}.

\bibitem[Paria et~al.(2017)Paria, Lahiri, and Biswas]{Biswajit2017}
Biswajit Paria, Avisek Lahiri, and Prabir~Kumar Biswas.
\newblock Policygan: Training generative adversarial networks using policy gradient.
\newblock In \emph{2017 Ninth International Conference on Advances in Pattern Recognition (ICAPR)}, pp.\  1--6, 2017.
\newblock \doi{10.1109/ICAPR.2017.8593063}.

\bibitem[Paszke et~al.(2019)Paszke, Gross, Massa, Lerer, Bradbury, Chanan, Killeen, Lin, Gimelshein, Antiga, Desmaison, K\"{o}pf, Yang, DeVito, Raison, Tejani, Chilamkurthy, Steiner, Fang, Bai, and Chintala]{pytorch}
Adam Paszke, Sam Gross, Francisco Massa, Adam Lerer, James Bradbury, Gregory Chanan, Trevor Killeen, Zeming Lin, Natalia Gimelshein, Luca Antiga, Alban Desmaison, Andreas K\"{o}pf, Edward Yang, Zach DeVito, Martin Raison, Alykhan Tejani, Sasank Chilamkurthy, Benoit Steiner, Lu~Fang, Junjie Bai, and Soumith Chintala.
\newblock \emph{PyTorch: an imperative style, high-performance deep learning library}.
\newblock Curran Associates Inc., Red Hook, NY, USA, 2019.

\bibitem[Pedregosa(2016)]{pedregosa2016}
Fabian Pedregosa.
\newblock Hyperparameter optimization with approximate gradient.
\newblock In Maria~Florina Balcan and Kilian~Q. Weinberger (eds.), \emph{Proceedings of The 33rd International Conference on Machine Learning}, volume~48 of \emph{Proceedings of Machine Learning Research}, pp.\  737--746, New York, New York, USA, 20--22 Jun 2016. PMLR.
\newblock URL \url{https://proceedings.mlr.press/v48/pedregosa16.html}.

\bibitem[Petrulionyte et~al.(2024)Petrulionyte, Mairal, and Arbel]{petrulionyte2024}
Ieva Petrulionyte, Julien Mairal, and Michael Arbel.
\newblock Functional bilevel optimization for machine learning.
\newblock In \emph{NeurIPS (Spotlight Poster)}, 2024.
\newblock arXiv preprint arXiv:2403.20233.

\bibitem[Radford \& Narasimhan(2018)Radford and Narasimhan]{Radford2018ImprovingLU}
Alec Radford and Karthik Narasimhan.
\newblock Improving language understanding by generative pre-training.
\newblock 2018.
\newblock URL \url{https://api.semanticscholar.org/CorpusID:49313245}.

\bibitem[Radford et~al.(2019)Radford, Wu, Child, Luan, Amodei, and Sutskever]{Radford2019LanguageMA}
Alec Radford, Jeff Wu, Rewon Child, David Luan, Dario Amodei, and Ilya Sutskever.
\newblock Language models are unsupervised multitask learners.
\newblock 2019.
\newblock URL \url{https://api.semanticscholar.org/CorpusID:160025533}.

\bibitem[Rafailov et~al.(2023)Rafailov, Sharma, Mitchell, Manning, Ermon, and Finn]{rafailov2023direct}
Rafael Rafailov, Archit Sharma, Eric Mitchell, Christopher~D Manning, Stefano Ermon, and Chelsea Finn.
\newblock Direct preference optimization: Your language model is secretly a reward model.
\newblock In \emph{Thirty-seventh Conference on Neural Information Processing Systems}, 2023.
\newblock URL \url{https://openreview.net/forum?id=HPuSIXJaa9}.

\bibitem[Rajeswaran et~al.(2019)Rajeswaran, Finn, Kakade, and Levine]{rajeswaran2019metalearningimplicitgradients}
Aravind Rajeswaran, Chelsea Finn, Sham Kakade, and Sergey Levine.
\newblock Meta-learning with implicit gradients, 2019.
\newblock URL \url{https://arxiv.org/abs/1909.04630}.

\bibitem[Ramesh et~al.(2021)Ramesh, Pavlov, Goh, Gray, Voss, Radford, Chen, and Sutskever]{ramesh2021zero}
Aditya Ramesh, Mikhail Pavlov, Gabriel Goh, Scott Gray, Chelsea Voss, Alec Radford, Mark Chen, and Ilya Sutskever.
\newblock Zero-shot text-to-image generation.
\newblock In Marina Meila and Tong Zhang (eds.), \emph{Proceedings of the 38th International Conference on Machine Learning}, volume 139 of \emph{Proceedings of Machine Learning Research}, pp.\  8821--8831. PMLR, 18--24 Jul 2021.
\newblock URL \url{https://proceedings.mlr.press/v139/ramesh21a.html}.

\bibitem[Ranzato et~al.(2016)Ranzato, Chopra, Auli, and Zaremba]{ranzato2016}
Marc'Aurelio Ranzato, Sumit Chopra, Michael Auli, and Wojciech Zaremba.
\newblock Sequence level training with recurrent neural networks, 2016.
\newblock URL \url{https://arxiv.org/abs/1511.06732}.

\bibitem[Ren et~al.(2023)Ren, Feng, Liu, Pan, Fu, Mai, and Yang]{JMLR:TorchOpt}
Jie Ren, Xidong Feng, Bo~Liu, Xuehai Pan, Yao Fu, Luo Mai, and Yaodong Yang.
\newblock Torchopt: An efficient library for differentiable optimization.
\newblock \emph{Journal of Machine Learning Research}, 24\penalty0 (367):\penalty0 1--14, 2023.
\newblock URL \url{http://jmlr.org/papers/v24/23-0191.html}.

\bibitem[Rombach et~al.(2021)Rombach, Blattmann, Lorenz, Esser, and Ommer]{rombach2021highresolution}
Robin Rombach, Andreas Blattmann, Dominik Lorenz, Patrick Esser, and Björn Ommer.
\newblock High-resolution image synthesis with latent diffusion models, 2021.

\bibitem[Shao et~al.(2024)Shao, Wang, Zhu, Xu, Song, Bi, Zhang, Zhang, Li, Wu, and Guo]{shao2024}
Zhihong Shao, Peiyi Wang, Qihao Zhu, Runxin Xu, Junxiao Song, Xiao Bi, Haowei Zhang, Mingchuan Zhang, Y.~K. Li, Y.~Wu, and Daya Guo.
\newblock Deepseekmath: Pushing the limits of mathematical reasoning in open language models, 2024.
\newblock URL \url{https://arxiv.org/abs/2402.03300}.

\bibitem[Shen et~al.(2024)Shen, Yang, and Chen]{shen2024}
Han Shen, Zhuoran Yang, and Tianyi Chen.
\newblock Principled penalty-based methods for bilevel reinforcement learning and rlhf, 2024.
\newblock URL \url{https://arxiv.org/abs/2402.06886}.

\bibitem[Shenfeld et~al.(2025)Shenfeld, Pari, and Agrawal]{shenfeld2025}
Idan Shenfeld, Jyothish Pari, and Pulkit Agrawal.
\newblock Rl's razor: Why online reinforcement learning forgets less, 2025.
\newblock URL \url{https://arxiv.org/abs/2509.04259}.

\bibitem[Shi et~al.(2025)Shi, Ichter, Equi, Ke, Pertsch, Vuong, Tanner, Walling, Wang, Fusai, Li-Bell, Driess, Groom, Levine, and Finn]{shi2025}
Lucy~Xiaoyang Shi, Brian Ichter, Michael Equi, Liyiming Ke, Karl Pertsch, Quan Vuong, James Tanner, Anna Walling, Haohuan Wang, Niccolo Fusai, Adrian Li-Bell, Danny Driess, Lachy Groom, Sergey Levine, and Chelsea Finn.
\newblock Hi robot: Open-ended instruction following with hierarchical vision-language-action models, 2025.
\newblock URL \url{https://arxiv.org/abs/2502.19417}.

\bibitem[Sohl-Dickstein et~al.(2015)Sohl-Dickstein, Weiss, Maheswaranathan, and Ganguli]{sohldickstein2015}
Jascha Sohl-Dickstein, Eric~A. Weiss, Niru Maheswaranathan, and Surya Ganguli.
\newblock Deep unsupervised learning using nonequilibrium thermodynamics, 2015.
\newblock URL \url{https://arxiv.org/abs/1503.03585}.

\bibitem[Song et~al.(2024)Song, Yu, Li, Yu, Huang, Li, and Wang]{song2024preference}
Feifan Song, Bowen Yu, Minghao Li, Haiyang Yu, Fei Huang, Yongbin Li, and Houfeng Wang.
\newblock Preference ranking optimization for human alignment, 2024.
\newblock URL \url{https://arxiv.org/abs/2306.17492}.

\bibitem[Song et~al.(2020)Song, Gao, Yang, Choromanski, Pacchiano, and Tang]{Song2020}
Xingyou Song, Wenbo Gao, Yuxiang Yang, Krzysztof Choromanski, Aldo Pacchiano, and Yunhao Tang.
\newblock Es-maml: Simple hessian-free meta learning.
\newblock In \emph{International Conference on Learning Representations}, 2020.
\newblock URL \url{https://openreview.net/forum?id=S1exA2NtDB}.

\bibitem[Sontakke et~al.(2023)Sontakke, Zhang, Arnold, Pertsch, Biyik, Sadigh, Finn, and Itti]{sontakke2023roboclip}
Sumedh~Anand Sontakke, Jesse Zhang, S{\'e}b Arnold, Karl Pertsch, Erdem Biyik, Dorsa Sadigh, Chelsea Finn, and Laurent Itti.
\newblock Robo{CLIP}: One demonstration is enough to learn robot policies.
\newblock In \emph{Thirty-seventh Conference on Neural Information Processing Systems}, 2023.
\newblock URL \url{https://openreview.net/forum?id=DVlawv2rSI}.

\bibitem[Stiennon et~al.(2020)Stiennon, Ouyang, Wu, Ziegler, Lowe, Voss, Radford, Amodei, and Christiano]{stiennon2020}
Nisan Stiennon, Long Ouyang, Jeff Wu, Daniel~M. Ziegler, Ryan Lowe, Chelsea Voss, Alec Radford, Dario Amodei, and Paul Christiano.
\newblock Learning to summarize from human feedback.
\newblock In \emph{Proceedings of the 34th International Conference on Neural Information Processing Systems}, NIPS '20, Red Hook, NY, USA, 2020. Curran Associates Inc.
\newblock ISBN 9781713829546.

\bibitem[Swamy et~al.(2025)Swamy, Choudhury, Sun, Wu, and Bagnell]{swamy2025}
Gokul Swamy, Sanjiban Choudhury, Wen Sun, Zhiwei~Steven Wu, and J.~Andrew Bagnell.
\newblock All roads lead to likelihood: The value of reinforcement learning in fine-tuning, 2025.
\newblock URL \url{https://arxiv.org/abs/2503.01067}.

\bibitem[Tan et~al.(2019)Tan, Hu, Yang, Salakhutdinov, and Xing]{tan2019connectingdotsmlerl}
Bowen Tan, Zhiting Hu, Zichao Yang, Ruslan Salakhutdinov, and Eric Xing.
\newblock Connecting the dots between mle and rl for sequence prediction, 2019.
\newblock URL \url{https://arxiv.org/abs/1811.09740}.

\bibitem[Touvron et~al.(2023)Touvron, Martin, Stone, Albert, Almahairi, Babaei, Bashlykov, Batra, Bhargava, Bhosale, Bikel, Blecher, Ferrer, Chen, Cucurull, Esiobu, Fernandes, Fu, Fu, Fuller, Gao, Goswami, Goyal, Hartshorn, Hosseini, Hou, Inan, Kardas, Kerkez, Khabsa, Kloumann, Korenev, Koura, Lachaux, Lavril, Lee, Liskovich, Lu, Mao, Martinet, Mihaylov, Mishra, Molybog, Nie, Poulton, Reizenstein, Rungta, Saladi, Schelten, Silva, Smith, Subramanian, Tan, Tang, Taylor, Williams, Kuan, Xu, Yan, Zarov, Zhang, Fan, Kambadur, Narang, Rodriguez, Stojnic, Edunov, and Scialom]{touvron2023llama2}
Hugo Touvron, Louis Martin, Kevin Stone, Peter Albert, Amjad Almahairi, Yasmine Babaei, Nikolay Bashlykov, Soumya Batra, Prajjwal Bhargava, Shruti Bhosale, Dan Bikel, Lukas Blecher, Cristian~Canton Ferrer, Moya Chen, Guillem Cucurull, David Esiobu, Jude Fernandes, Jeremy Fu, Wenyin Fu, Brian Fuller, Cynthia Gao, Vedanuj Goswami, Naman Goyal, Anthony Hartshorn, Saghar Hosseini, Rui Hou, Hakan Inan, Marcin Kardas, Viktor Kerkez, Madian Khabsa, Isabel Kloumann, Artem Korenev, Punit~Singh Koura, Marie-Anne Lachaux, Thibaut Lavril, Jenya Lee, Diana Liskovich, Yinghai Lu, Yuning Mao, Xavier Martinet, Todor Mihaylov, Pushkar Mishra, Igor Molybog, Yixin Nie, Andrew Poulton, Jeremy Reizenstein, Rashi Rungta, Kalyan Saladi, Alan Schelten, Ruan Silva, Eric~Michael Smith, Ranjan Subramanian, Xiaoqing~Ellen Tan, Binh Tang, Ross Taylor, Adina Williams, Jian~Xiang Kuan, Puxin Xu, Zheng Yan, Iliyan Zarov, Yuchen Zhang, Angela Fan, Melanie Kambadur, Sharan Narang, Aurelien Rodriguez, Robert Stojnic, Sergey Edunov, and Thomas
  Scialom.
\newblock Llama 2: Open foundation and fine-tuned chat models, 2023.
\newblock URL \url{https://arxiv.org/abs/2307.09288}.

\bibitem[Uehara et~al.(2024)Uehara, Zhao, Hajiramezanali, Scalia, Eraslan, Lal, Levine, and Biancalani]{uehara2024}
Masatoshi Uehara, Yulai Zhao, Ehsan Hajiramezanali, Gabriele Scalia, Gökcen Eraslan, Avantika Lal, Sergey Levine, and Tommaso Biancalani.
\newblock Bridging model-based optimization and generative modeling via conservative fine-tuning of diffusion models, 2024.
\newblock URL \url{https://arxiv.org/abs/2405.19673}.

\bibitem[Van~Rossum \& Drake~Jr(1995)Van~Rossum and Drake~Jr]{van1995python}
Guido Van~Rossum and Fred~L Drake~Jr.
\newblock \emph{Python reference manual}.
\newblock Centrum voor Wiskunde en Informatica Amsterdam, 1995.

\bibitem[Vaswani et~al.(2023)Vaswani, Shazeer, Parmar, Uszkoreit, Jones, Gomez, Kaiser, and Polosukhin]{vaswani2023attentionneed}
Ashish Vaswani, Noam Shazeer, Niki Parmar, Jakob Uszkoreit, Llion Jones, Aidan~N. Gomez, Lukasz Kaiser, and Illia Polosukhin.
\newblock Attention is all you need, 2023.
\newblock URL \url{https://arxiv.org/abs/1706.03762}.

\bibitem[Venkatraman et~al.(2015)Venkatraman, Hebert, and Bagnell]{venkatraman2015}
Arun Venkatraman, Martial Hebert, and J.~Andrew Bagnell.
\newblock Improving multi-step prediction of learned time series models.
\newblock In \emph{Proceedings of the Twenty-Ninth AAAI Conference on Artificial Intelligence}, AAAI'15, pp.\  3024–3030. AAAI Press, 2015.
\newblock ISBN 0262511290.

\bibitem[Volkovs et~al.(2011)Volkovs, Larochelle, and Zemel]{volkovs2011}
Maksims~N. Volkovs, Hugo Larochelle, and Richard~S. Zemel.
\newblock Loss-sensitive training of probabilistic conditional random fields, 2011.
\newblock URL \url{https://arxiv.org/abs/1107.1805}.

\bibitem[von Stackelberg(1934)]{stackelberg1934}
Heinrich von Stackelberg.
\newblock \emph{Marktform und Gleichgewicht}.
\newblock Springer-Verlag, Berlin, 1934.
\newblock English translation: \emph{The Theory of the Market Economy}, Oxford University Press, 1952.

\bibitem[Wang(2023)]{uci}
Jingcong Wang.
\newblock Uci datasets, 2023.
\newblock URL \url{https://dx.doi.org/10.21227/g4y0-sw34}.

\bibitem[Wang et~al.(2025)Wang, Yu, Wan, Gan, and Zhan]{wang2025founder}
Yucen Wang, Rui Yu, Shenghua Wan, Le~Gan, and De-Chuan Zhan.
\newblock {FOUNDER}: Grounding foundation models in world models for open-ended embodied decision making.
\newblock In \emph{Forty-second International Conference on Machine Learning}, 2025.
\newblock URL \url{https://openreview.net/forum?id=UTT5OTyIWm}.

\bibitem[Wen et~al.(2024)Wen, Liao, Deng, Wang, Zhang, and Wen]{wen2024entropy}
Muning Wen, Junwei Liao, Cheng Deng, Jun Wang, Weinan Zhang, and Ying Wen.
\newblock Entropy-regularized token-level policy optimization for language agent reinforcement, 2024.
\newblock URL \url{https://arxiv.org/abs/2402.06700}.

\bibitem[Wengert(1964)]{wengert1964}
R.~E. Wengert.
\newblock A simple automatic derivative evaluation program.
\newblock \emph{Commun. ACM}, 7\penalty0 (8):\penalty0 463–464, August 1964.
\newblock ISSN 0001-0782.
\newblock \doi{10.1145/355586.364791}.
\newblock URL \url{https://doi.org/10.1145/355586.364791}.

\bibitem[Williams(1992)]{Williams1992}
R.~J. Williams.
\newblock Simple statistical gradient-following algorithms for connectionist reinforcement learning.
\newblock \emph{Machine Learning}, 8:\penalty0 229--256, 1992.

\bibitem[Xiao et~al.(2025)Xiao, Yuan, Saif, Liu, Kompella, Wang, and Chen]{xiao2025a}
Quan Xiao, Hui Yuan, A~F~M Saif, Gaowen Liu, Ramana~Rao Kompella, Mengdi Wang, and Tianyi Chen.
\newblock A first-order generative bilevel optimization framework for diffusion models.
\newblock In \emph{Forty-second International Conference on Machine Learning}, 2025.
\newblock URL \url{https://openreview.net/forum?id=3qL4LRUaJ8}.

\bibitem[Xin et~al.(2025)Xin, Ren, Song, Shao, Zhao, Wang, Liu, Zhang, Lu, Du, Gao, Zhang, Zhu, Yang, Gou, Wu, Luo, and Ruan]{xin2025deepseekproverv}
Huajian Xin, Z.Z. Ren, Junxiao Song, Zhihong Shao, Wanjia Zhao, Haocheng Wang, Bo~Liu, Liyue Zhang, Xuan Lu, Qiushi Du, Wenjun Gao, Haowei Zhang, Qihao Zhu, Dejian Yang, Zhibin Gou, Z.F. Wu, Fuli Luo, and Chong Ruan.
\newblock Deepseek-prover-v1.5: Harnessing proof assistant feedback for reinforcement learning and monte-carlo tree search.
\newblock In \emph{The Thirteenth International Conference on Learning Representations}, 2025.
\newblock URL \url{https://openreview.net/forum?id=I4YAIwrsXa}.

\bibitem[Yang et~al.(2025)Yang, Gao, and xiang Yuan]{yang2025}
Yan Yang, Bin Gao, and Ya~xiang Yuan.
\newblock Bilevel reinforcement learning via the development of hyper-gradient without lower-level convexity, 2025.
\newblock URL \url{https://arxiv.org/abs/2405.19697}.

\bibitem[Yin et~al.(2024)Yin, Fu, Zhao, Li, Sun, Xu, and Chen]{Yin2024}
Shukang Yin, Chaoyou Fu, Sirui Zhao, Ke~Li, Xing Sun, Tong Xu, and Enhong Chen.
\newblock A survey on multimodal large language models.
\newblock \emph{National Science Review}, 11\penalty0 (12), November 2024.
\newblock ISSN 2053-714X.
\newblock \doi{10.1093/nsr/nwae403}.
\newblock URL \url{http://dx.doi.org/10.1093/nsr/nwae403}.

\bibitem[Younis et~al.(2024)Younis, Perez-Vicente, Balis, Dudley, Davey, and Terry]{minari}
Omar~G. Younis, Rodrigo Perez-Vicente, John~U. Balis, Will Dudley, Alex Davey, and Jordan~K Terry.
\newblock Minari, September 2024.
\newblock URL \url{https://doi.org/10.5281/zenodo.13767625}.

\bibitem[Yu et~al.(2017)Yu, Zhang, Wang, and Yu]{yu2017seqgan}
Lantao Yu, Weinan Zhang, Jun Wang, and Yong Yu.
\newblock Seqgan: Sequence generative adversarial nets with policy gradient, 2017.
\newblock URL \url{https://arxiv.org/abs/1609.05473}.

\bibitem[Yu et~al.(2020)Yu, Thomas, Yu, Ermon, Zou, Levine, Finn, and Ma]{Yu2020}
Tianhe Yu, Garrett Thomas, Lantao Yu, Stefano Ermon, James~Y Zou, Sergey Levine, Chelsea Finn, and Tengyu Ma.
\newblock Mopo: Model-based offline policy optimization.
\newblock In H.~Larochelle, M.~Ranzato, R.~Hadsell, M.F. Balcan, and H.~Lin (eds.), \emph{Advances in Neural Information Processing Systems}, volume~33, pp.\  14129--14142. Curran Associates, Inc., 2020.
\newblock URL \url{https://proceedings.neurips.cc/paper/2020/file/a322852ce0df73e204b7e67cbbef0d0a-Paper.pdf}.

\bibitem[Zekri \& Boullé(2025)Zekri and Boullé]{zekri2025}
Oussama Zekri and Nicolas Boullé.
\newblock Fine-tuning discrete diffusion models with policy gradient methods, 2025.
\newblock URL \url{https://arxiv.org/abs/2502.01384}.

\bibitem[Zeng et~al.(2022)Zeng, Li, Garcia, and Hong]{zeng2022maximum}
Siliang Zeng, Chenliang Li, Alfredo Garcia, and Mingyi Hong.
\newblock Maximum-likelihood inverse reinforcement learning with finite-time guarantees.
\newblock \emph{Advances in Neural Information Processing Systems}, 35:\penalty0 10122--10135, 2022.

\bibitem[Zhang et~al.(2024)Zhang, Huang, Jin, and Lu]{zhang2024}
Jingyi Zhang, Jiaxing Huang, Sheng Jin, and Shijian Lu.
\newblock Vision-language models for vision tasks: A survey, 2024.
\newblock URL \url{https://arxiv.org/abs/2304.00685}.

\bibitem[Ziebart et~al.(2008)Ziebart, Maas, Bagnell, and Dey]{ziebart2008}
Brian~D. Ziebart, Andrew Maas, J.~Andrew Bagnell, and Anind~K. Dey.
\newblock Maximum entropy inverse reinforcement learning.
\newblock In \emph{Proceedings of the 23rd National Conference on Artificial Intelligence - Volume 3}, AAAI'08, pp.\  1433–1438. AAAI Press, 2008.
\newblock ISBN 9781577353683.

\bibitem[Zou et~al.(2019)Zou, Ren, Yan, Su, and Zhu]{zou2019reward}
Haosheng Zou, Tongzheng Ren, Dong Yan, Hang Su, and Jun Zhu.
\newblock Reward shaping via meta-learning.
\newblock \emph{arXiv preprint arXiv:1901.09330}, 2019.

\bibitem[Łajszczak et~al.(2024)Łajszczak, Cámbara, Li, Beyhan, van Korlaar, Yang, Joly, Álvaro Martín-Cortinas, Abbas, Michalski, Moinet, Karlapati, Muszyńska, Guo, Putrycz, Gambino, Yoo, Sokolova, and Drugman]{lajszczak2024}
Mateusz Łajszczak, Guillermo Cámbara, Yang Li, Fatih Beyhan, Arent van Korlaar, Fan Yang, Arnaud Joly, Álvaro Martín-Cortinas, Ammar Abbas, Adam Michalski, Alexis Moinet, Sri Karlapati, Ewa Muszyńska, Haohan Guo, Bartosz Putrycz, Soledad~López Gambino, Kayeon Yoo, Elena Sokolova, and Thomas Drugman.
\newblock Base tts: Lessons from building a billion-parameter text-to-speech model on 100k hours of data, 2024.
\newblock URL \url{https://arxiv.org/abs/2402.08093}.

\end{thebibliography}
\bibliographystyle{main}


\newpage
\appendix

\textbf{\LARGE Appendix}


\addtocontents{toc}{\protect\setcounter{tocdepth}{3}}

\renewcommand*\contentsname{\Large Table of Contents}

\tableofcontents
\clearpage

\section{Theoretical analysis}
\label{app:theory}

\subsection{Proof of \cref{prop:outer}}
\label{app:inner}

\subsubsection{A useful lemma}

This lemma will be useful to show to concavity of the objective function in the following proposition. 

\begin{boxlem}\label{positive}
Let $(A,B,C)\in \mathrm{S}^{+}_n\times \mathrm{S}^{+}_n \times \mathbb{R}^{n\times n}$, then one has that:
\[
\operatorname{Tr}(A C B C^\top) \geq 0.
\]
\end{boxlem}
\begin{proof}
Since \( A \) is symmetric positive semidefinite, there exists a symmetric matrix \( A^{1/2} \) such that
\[
A = (A^{1/2})^2.
\]
Then,
\[
\operatorname{Tr}(A C B C^\top) = \operatorname{Tr}(A^{1/2} A^{1/2} C B C^\top).
\] 
Then, it follows that : 
\[
\operatorname{Tr}(A^{1/2} A^{1/2} C B C^\top) = \operatorname{Tr}(A^{1/2} C B C^\top A^{1/2}).
\]
Let
\[
M = A^{1/2} C B C^\top A^{1/2}.
\]
Then,
\[
\operatorname{Tr}(A C B C^\top) = \operatorname{Tr}(M).
\]
So now it suffices to show that $M$ is definite semipositive.

The matrix \( M \) is symmetric. For any vector \( x\in \RR^{n} \),
\[
x^\top M x = x^\top A^{1/2} C B C^\top A^{1/2} x = (C^\top A^{1/2} x)^\top B (C^\top A^{1/2} x).
\]
Since \( B \) is positive semidefinite,
\[
(C^\top A^{1/2} x)^\top B (C^\top A^{1/2}x) \geq 0.
\]
Hence, \( M \) is positive semidefinite.
\end{proof}

\subsubsection{An intermediate proposition: solution of the inner-level problem}

We start by proving the following proposition on the closed-form solution of the inner level problem in \ref{eq:bo}:

\begin{boxprop}\label{prop:inner}
Under Assumptions \ref{ass:model_general} and \ref{ass:reward}, the inner-level optimization problem
\begin{equation*}\label{low}
\theta^\star_U = \argmax_{\theta \in \Theta} \Exp{X,Y \sim q}{
\Exp{\widehat{Y} \mid X \sim \hat{p}_{\theta}}{
-(\widehat{Y} - Y)^T U (\widehat{Y} - Y)} + \lambda \mathcal{H}(\hat{p}_{\theta})
}. 
\end{equation*}
admits exactly one solution that writes as \[
\boxed{ \;
\theta^{\star}(U) \;=\;
\left( \Lambda, \; \frac{\lambda U^{-1}}{2} \right).
\;}
\]
\end{boxprop}

\begin{proof}[Proof of Proposition~\ref{prop:inner}]

   We prove the proposition by deriving a closed-form expression for the objective $\theta \mapsto J(\theta)$ and it's gradient, then we show that $J$ is strictly concave in $\theta=(A,B)$ over $\mathbb{R}^{n\times n}\times S_{n}^{++}(\mathbb{R})$ which guarantee that there is exactly one  solution.

The objective function is:
\[
J(\theta) = \mathbb{E}_{X} \mathbb{E}_{Y \mid X} \left[ \mathbb{E}_{\widehat{Y} \sim P_{\theta}} \left[ -(\widehat{Y} - Y)^T U (\widehat{Y} - Y) + \lambda H(\widehat{Y} \mid X) \right] \right].
\]
First, we compute the inner expectation over $\widehat{Y}$ for fixed $X$ and $Y$. Since $\widehat{Y} \mid X \sim \mathcal{N}(AX, B)$, the entropy of $\widehat{Y} \mid X$ is:
\[
H(\widehat{Y} \mid X) = \frac{1}{2} \log(2\pi e \operatorname{det}(B)).
\]
Now, define:
\[
I_{\widehat{Y}}: = \mathbb{E}_{\widehat{Y} \sim P_{\theta}} \left[ -(\widehat{Y} - Y)^T U (\widehat{Y} - Y) + \lambda H(\widehat{Y} \mid X) \right].
\]

\[
I_{\widehat{Y}} = \underbrace{\mathbb{E}_{\widehat{Y} \sim P_{\theta}} \left[ -(\widehat{Y} - Y)^T U (\widehat{Y} - Y) \right]}_{ \mathcal{A}} + \frac{\lambda}{2} \log(2\pi e \operatorname{det}(B)).
\]

For fixed $X$ and $Y$, let $Z = \widehat{Y} - Y$. We show that :
\[
\mathcal{A} = - \left[ X^T A^T U A X +  \operatorname{Tr}(UB) - 2Y^T U (AX) + Y^T U Y \right].
\]
Expanding the quadratic form:
\[
Z^T U Z = \widehat{Y}^T U \widehat{Y} - 2Y^T U \widehat{Y} + Y^T U Y.
\]
Taking expectations:
\[
\mathbb{E}_{\widehat{Y} \mid X} \left[ \widehat{Y}^T U \widehat{Y} - 2Y^T U \widehat{Y} + Y^T U Y \right] = \mathbb{E}[\widehat{Y}^T U \widehat{Y}] - 2Y^T U \mathbb{E}[\widehat{Y}] + Y^T U Y.
\]
Since $\widehat{Y} \mid X \sim \mathcal{N}(AX, B)$, we have $\mathbb{E}[\widehat{Y} \mid X] = AX$. Using the formula for the expectation of a quadratic form, for a random vector $W$ with mean $\mu$ and covariance $K$:
\[
\mathbb{E}[W^T U W] = \mu^T U \mu + \operatorname{Tr}(U K).
\]
Here, $W = \widehat{Y}$, $\mu = AX$, $K = B$, so:
\[
\mathbb{E}[\widehat{Y}^T U \widehat{Y} \mid X] = (AX)^T U (AX) + \operatorname{Tr}(U B).
\]
Thus, we get the desired expression for $\mathcal{A}$.
It follows that,
\[
I_{\widehat{Y}} = -X^T A^T U A X -  \operatorname{Tr}(UB) + 2Y^T U A X - Y^T U Y + \frac{\lambda}{2} \log(2\pi e \operatorname{det}(B)).
\]

Now, for fixed $X$, we compute:
\[
J_X(A, B) = \mathbb{E}_{Y \mid X} \left[ I_{\widehat{Y}} \right].
\]
Since $Y \mid X \sim \mathcal{N}(\Lambda X, \Sigma)$, we have $\mathbb{E}[Y \mid X] = \Lambda X$. Using  quadratic form expectation again:
\[
\mathbb{E}_{Y \mid X} [Y^T U Y] =X^T \Lambda^T U \Lambda X + \operatorname{Tr}(U \Sigma).
\]
Thus,
\begin{align*}
J_X(A, B) &= \mathbb{E}_{Y \mid X} \left[ -X^T A^T U A X -  \operatorname{Tr}(UB) + 2Y^T U A X - Y^T U Y \right] + \frac{\lambda}{2} \log(2\pi e \operatorname{det}(B)) \\
&= -X^T A^T U A X -  \operatorname{Tr}(UB) + 2 \mathbb{E}_{Y \mid X}[Y]^T U A X - \mathbb{E}_{Y \mid X}[Y^T U Y] + \frac{\lambda}{2} \log(2\pi e \operatorname{det}(B))\\
&= -X^T A^T U A X - \operatorname{Tr}(UB) + 2 (\Lambda X)^T U A X - \left( X^T \Lambda^T U \Lambda X +  \operatorname{Tr}(U\Sigma) \right) + \frac{\lambda}{2} \log(2\pi e \operatorname{det}(B))\\ 
&= -X^T \left( A^T U A - 2 \Lambda^T U A + \Lambda^T U \Lambda \right) X -  \operatorname{Tr}(U(B+ \Sigma)) + \frac{\lambda}{2} \log(2\pi e \operatorname{det}(B)).
\end{align*}
Since $U$ is symmetric:
\[
A^T U A - 2 \Lambda^T U A + \Lambda^T U \Lambda = (A - \Lambda)^T U (A - \Lambda),
\]
 One has that:

\[
J(\theta)= \mathbb{E}_X \left[ -X^T (A - \Lambda)^T U (A - \Lambda) X \right] - \operatorname{Tr}(U(B+ \Sigma)) +  \frac{\lambda}{2} \log(2\pi e \operatorname{det}(B))
\]
Let $M_A = (A - \Lambda)^T U (A - \Lambda)$. it follows that (since $\operatorname{Tr}(x)=x$ for any $x\in \mathbb{R}$ and here $X^T M_A X\in \mathbb{R}$):
\[
\mathbb{E}_X \left[ X^T M_A X \right] = \mathbb{E}_X \left[ \operatorname{Tr}(X^T M_A X) \right] = \mathbb{E}_X \left[ \operatorname{Tr}(M_A X X^T) \right] = \operatorname{Tr}(M_A \mathbb{E}_X [X X^T]).
\]
Let $\Sigma_X = \mathbb{E}_X [X X^T]$. Thus,
\[
\mathbb{E}_X \left[ X^T M_A X \right] = \operatorname{Tr}\left( (A - \Lambda)^T U (A - \Lambda) \Sigma_X \right).
\]

\begin{equation}\label{eq:closed_form}
J(A,B) = -\operatorname{Tr}\left( (A - \Lambda)^T U (A - \Lambda) \Sigma_X \right) - \operatorname{Tr}(U(B+ \Sigma)) +  \frac{\lambda}{2} \log(2\pi e \operatorname{det}(B))
\end{equation}

Let $t\in \mathbb{R}$ and $u_{1}=(A_{1},B_{1})$ and $u_{2}=(A_{2},B_2)$ such that $u_1+tu_2\in \mathbb{R}^{n\times n}\times S_{n}^{++}$.
It suffices to show that 
$g : t\mapsto J(u_{1}+tu_{2})$ is a concave function on $N=\left\{ t\in \mathbb{R}, \quad u_{1}+tu_{2}\in \mathbb{R}^{n\times n}\times S_{n}^{++}\right\}$.

Let $t\in N$, one has $$u_{1}+tu_{2}=(A_{1}+tA_2,B_1+tB_2),$$
\begin{align*}
g(t) &= 
- \underbrace{ \operatorname{Tr}(U\left(B_1+tB_2+U\Sigma\right) -\operatorname{Tr}\!\left( (A_{1} + tA_{2} - \Lambda)^{\!\top} 
\, U \, (A_{1} + tA_{2} - \Lambda) \, \Sigma_X \right)}_{:=H_{1}(t)} 
\\
&\quad + \underbrace{\frac{\lambda}{2} \log\left(2\pi e \operatorname{det}(B_1+tB_2)\right)}_{:=H_{2}(t)} .
\end{align*}
First, $g\in C^{2}\left(N,\mathbb{R}\right)$.

Regarding $H_1$, a straightforward calculation shows that 

\[
t\mapsto H_1(t) = \alpha t^2 + \beta t + \gamma,
\]
where:
\begin{align*}
\alpha &= -\operatorname{Tr}(U A_2 \Sigma_X A_2^\top), \\
\beta &= -2 \operatorname{Tr}(U (A_1-\Lambda) \Sigma_X A_2^\top)-\operatorname{Tr}(UB_2), \\
\gamma &= -\operatorname{Tr}(U (A_1 -\Lambda) \Sigma_X (A_1-\Lambda)^\top)-\operatorname{Tr}(UB_1+U^2\Sigma).
\end{align*}
The second derivative is:
\[
t\mapsto H_1''(t)= -2 \operatorname{Tr}(U A_2 \Sigma_X A_2^\top).
\]
Since \( U, \Sigma_X \in S_{n}^{+}(\mathbb{R}) \), and $A_2\in \RR^{n\times n}$ we apply the lemma \ref{positive} which gives us immediately that 
 \(\operatorname{Tr}(U A_2 \Sigma_X A_2^\top) \geq 0\).
Thus: $H_1$ is concave.

For $H_2$ one can show, using Jacobi's formulas that
$$\forall t \in N\quad H^{\prime}_2(t)=\frac{d}{dt} \log(\det(B_1 + (\cdot) B_2)) = \operatorname{Tr}\left( (B_1 + t B_2)^{-1} B_2 \right)$$

Since $B_1,B_2\in S_{n}^{++}$ on can find two basis $\mathcal{B}_1$ and $\mathcal{B}_2$ such that they are diagonal in these bases, 
\[
\begin{aligned}
&\exists \boldsymbol{\lambda} = (\lambda_1, \dots, \lambda_n) \in \mathbb{R}^n \setminus \{\mathbf{0}_n\}, \quad 
 \exists \boldsymbol{\mu} = (\mu_1, \dots, \mu_n) \in \mathbb{R}^n \setminus \{\mathbf{0}_n\} \\
&\text{such that} \quad 
(B_1)_{\mathcal{B}_1} = \operatorname{Diag}(\boldsymbol{\lambda}) \quad \text{and} \quad 
(B_2)_{\mathcal{B}_2} = \operatorname{Diag}(\boldsymbol{\mu})
\end{aligned}
\]
So 
$$\forall t \in N\quad H^{\prime}_{2}(t)=\sum_{1\leq i\leq n}\frac{\mu_i}{\lambda_i+t\mu_i},$$
so 
$$\forall t\in N\quad H^{\prime \prime}_{2}(t)=-\sum_{1\leq i\leq n}\frac{\mu_i^2}{(\lambda_i+t\mu_i)^2}<0.$$

So $g$ is a sum of a concave and a strictly concave function so it's a strictly concave function and thus $J$ is strictly concave, thus the problem \ref{low} admits exactly one solution on $ \mathbb{R}^{n\times n}\times S_{n}^{++}(\mathbb{R})$; which is solution of
\begin{equation}
    \nabla J (A,B)=0.
\end{equation}
Let's find in closed form the solutions of the previous equation.

It follows that:
\[
\nabla_A J(A,B) = -2U(A - \Lambda)\Sigma_X
\]

Then,

$$\nabla_{B} J(A,B)=-U^T+\frac{\lambda}{2} \frac{\partial \ln(\operatorname{det})(B)}{\partial B}=(B^{-1})^T=-U+\frac{\lambda}{2} B^{-1}$$
Finally:
\[
\boxed{ \;
\theta^{\star}(U) \;=\;
\left( \Lambda, \; \frac{\lambda U^{-1}}{2} \right).
\;}
\]

\end{proof}

\subsubsection{Concluding the proof: solution of the outer-level problem}
\label{app:outer}

We restate the proposition before proceeding with the proof:

\begin{boxprop}
Given the above assumptions and the solution in Proposition \ref{prop:inner}, the outer-level problem
\begin{equation*}
U^{\star} = \argmin_{U\in S_{n}^{++}(\RR)} \Exp{X,Y \sim q}{\log{\hat{p}_{\theta^{\star}_U} \mleft(Y|X\mright)}},
\end{equation*}
has exactly one solution:
\begin{equation*}
\boxed{U^{\star}=\frac{\lambda \Sigma^{-1}}{2}.}
\end{equation*}
\end{boxprop}

\begin{proof}
Let $U\in S_{n}^{++}(\mathbb{R})$ and $(\lambda,n) \in \mathbb{R}^{\star}_{+}\times \mathbb{N}^{\star}$ by the previous proposition one has 
$\theta^{\star}(U)=\left(\Lambda, \frac{\lambda U^{-1} }{2}\right)$.

Let's check that $$\varphi : U \mapsto \mathbb{E}_{X}D_{\mathrm{KL}}\bigl(q(\cdot \mid X) \,\|\, p_{\theta^{\star}_{U}}(\cdot \mid X)\bigr)$$ is a convex function of $U$.

To show the convexity of $\varphi$, we show the convexity of $g$, which is defined as follow.
Let $U,V\in S^{++}_{n}(\mathbb{R})$ and $t\in I_{U,V}:=\left\{ u\in \mathbb{R}\quad  U+uV\in S_{n}^{++}(\mathbb{R})\right\}$. 
Define $$\forall t\in I_{U,V}\quad g(t)=\varphi(tU+V).$$
One has that $g\in C^{2}\left(I_{U,V},\mathbb{R}\right)$.

Since both $q(\cdot \mid X)$ and $p_{\theta^{\star}_{U}}(\cdot \mid X)$ are Gaussian with the same mean $\Lambda X$, the Kullback–Leibler divergence has a closed-form expression:
\[
\text{D}_{\text{KL}}\bigl(q \,\|\, p_{\theta^{\star}_{U}}\bigr) = \frac{1}{2} \left[ \operatorname{Tr}\left( \Sigma_p^{-1} \Sigma \right) - n + \ln\left( \frac{\det(\Sigma_p)}{\det(\Sigma)} \right) \right],
\]
where $\Sigma_{p_{\theta^{\star}_{U}}} = \frac{\lambda}{2} U^{-1}$. It follows that,
\[
\Sigma_{p_{\theta^{\star}_{U}}}^{-1} = \frac{2}{\lambda} U, \quad \text{and} \quad \det(\Sigma_{p_{\theta^{\star}_{U}}}^{-1}) = \left( \frac{\lambda}{2} \right)^n \det(U)^{-1}.
\]
Substituting these in, we find:
\begin{align*}
\text{D}_{\text{KL}}\bigl(q \,\|\, p_{\theta^{\star}_{U}}\bigr) &= \frac{1}{2} \left[ \operatorname{Tr}\left( \frac{2}{\lambda} U \Sigma \right) - n + \ln\left( \frac{(\lambda/2)^n}{\det(U)\det(\Sigma)} \right) \right] \\
&= \frac{1}{\lambda} \operatorname{Tr}(U \Sigma) - \frac{n}{2} + \frac{n}{2} \ln\left( \frac{\lambda}{2} \right) - \frac{1}{2} \ln(\det(\Sigma)) - \frac{1}{2} \ln(\det(U)).
\end{align*}
This expression is independent of $X$, so its expectation is itself:
\[
\varphi(U) = \frac{1}{\lambda} \operatorname{Tr}(U \Sigma) - \frac{1}{2} \ln(\det(U)) + C,
\]
where $C$ is a constant independent of $U$.

Now, we express $g(t)$ explicitly and set $\forall t \in I_{U,V}\quad A(t)=U+tV$:
\[
\forall t\in I_{U,V}\quad g(t) = \varphi(A(t)) = \frac{1}{\lambda} \operatorname{Tr}(A(t) \Sigma) - \frac{1}{2} \ln(\det(A(t))) + C.
\]
Clearly, $g\in C^{2}(I_{U,V}, \mathbb{R})$, let's show that its second derivative is positive.

The first derivative is:
\[ \forall t\in I_{U,V},\quad
g'(t) = \frac{1}{\lambda} \operatorname{Tr}(U \Sigma) - \frac{1}{2} \frac{d}{dt} \ln(\det(A(t))).
\]
Using the identity that follows from Jacobi's formula: $$\forall t\in I_{U,V},\quad \frac{d}{dt} \ln(\det(A(t))) = \operatorname{Tr}\left( A(t)^{-1} A^{\prime}(t) \right),$$ we get:
\[\forall t\in I_{U,V},\quad
g'(t) = \frac{1}{\lambda} \operatorname{Tr}(U \Sigma) - \frac{1}{2} \operatorname{Tr}(A(t)^{-1} U).
\]

Differentiating again, one has that:
\[
\forall t\in I_{U,V},\quad g''(t) = -\frac{1}{2} \frac{d}{dt} \operatorname{Tr}(A(t)^{-1} U).
\]

Therefore,
\[\forall t\in I_{U,V},\quad
g''(t) = \frac{1}{2} \operatorname{Tr}(A(t)^{-1} U A(t)^{-1} U).
\]

Let $t\in I_{U,V}$ and $B = A(t)^{-1/2} U A(t)^{-1/2}$, where $A(t)^{1/2}$ is the symmetric positive definite square root of $A(t)$. Since $U$ is positive definite, $B$ is also positive definite. We have:
\begin{align*}
\operatorname{Tr}(A(t)^{-1} U A(t)^{-1} U) &= \operatorname{Tr}(A(t)^{-1/2} A(t)^{-1/2} U A(t)^{-1/2} A(t)^{-1/2} U) \\
&= \operatorname{Tr}(A(t)^{-1/2} U A(t)^{-1/2} A(t)^{-1/2} U A(t)^{-1/2}) \\
&= \operatorname{Tr}(B B) = \operatorname{Tr}(B^2).
\end{align*}
Thus,
\[
g''(t) = \frac{1}{2} \operatorname{Tr}(B^2).
\]
Since $B$ is symmetric and positive definite, its eigenvalues $\lambda_1, \dots, \lambda_n$ are positive. Therefore,
\[
\operatorname{Tr}(B^2) = \sum_{i=1}^n \lambda_i^2 > 0,
\]
which implies $g''(t) > 0$ for all $t \in I_{U,V}$.

Since the second derivative of $g$ is strictly positive on its domain, $g$ is strictly convex. Thus $\varphi$ is strictly convex on $S_{n}^{++}(\mathbb{R})$. 

The existence and the uniqueness is shown.

By the moment matching principle for Kullback–Leibler divergence one find that 
\[\boxed{U^{\star}=\frac{\lambda \Sigma^{-1}}{2}.}\]
\end{proof}

\subsection{Proof of \cref{cor:isotropic}}

\begin{boxcor}[Isotropic case]
If we assume that $\mathrm{B} = \sigma^2 \identity_n$, the set of solutions to the outer-level problem is characterized by:
\[\boxed{
\mathrm{U}^\star \in F_{\lambda,\Sigma}:=\left\{ \mathrm{U} \in S_n^{++}(\mathbb{R}) \, \middle| \, \operatorname{Tr}(\mathrm{U})= \frac{\lambda n^2}{2\operatorname{Tr}(\Sigma)} \right\}.}
\]
\end{boxcor}

\begin{proof}
    The proof follows the same calculations and arguments as those used in the proof of Proposition~\ref{prop:inner} and Proposition ~\ref{prop:outer}. 
    Specifically, we show that the objective function $(A, \sigma^2)\mapsto J(A, \sigma^2)$ is concave in $(A, \sigma^2)$ and solve the first-order optimality conditions. 
    This leads to the solution \[\theta^\star(U) = \left( \Lambda, \frac{\lambda n}{2\operatorname{Tr}(U)} \right).\]
    Substituting this solution into the $\text{D}_{KL}$ expression and following the same arguments leads to  $F_{\lambda,\Sigma}$.
\end{proof}

\subsection{Proof of \cref{cor:reverse_kl} }
\label{app:corr45}

We first restate the corollary on the reverse KL minimization equivalence:

\begin{boxcor}
\small
Under the assumptions of Proposition \ref{prop:outer}, the optimal parameters $\theta^\star_{\mathrm{U}^\star}$ obtained from the lower-level problem with $\mathrm{U}^\star = \frac{\lambda}{2}\Sigma^{-1}$ minimize the reverse KL divergence between  $\hat{p}_\theta$ and $q$:
\begin{center}
$\theta^\star_{\mathrm{U}^\star} = \argmin_{\theta \in \Theta} \mathbb{E}_{X} \left[ d_{KL} \left( \hat{p}_\theta(\cdot|X) \,\|\, q(\cdot|X) \right) \right].$
\end{center}
\end{boxcor}

\begin{proof}

Substituting $\mathrm{U}^\star = \frac{\lambda}{2}\Sigma^{-1}$ gives:
\[
J(\theta) = \mathbb{E}_{X}\Exp{Y\mid X \sim q} {\left[ \mathbb{E}_{\widehat{Y} \sim \hat{p}_\theta(\cdot|X)} \left[ -\frac{\lambda}{2} (\widehat{Y} - Y)^\top \Sigma^{-1} (\widehat{Y} - Y) \right] \right]} + \lambda \mathbb{H}(\hat{p}_\theta).
\]

 Since $q(Y|X)$ is Gaussian with mean $AX$ and covariance $\Sigma$, write $Y = AX + \varepsilon$ with $\varepsilon \sim \mathcal{N}(0, \Sigma)$. Remark that:
\[
\widehat{Y} - Y = (\widehat{Y} - AX) - \varepsilon.
\]
Thus,
\[
(\widehat{Y} - Y)^\top \Sigma^{-1} (\widehat{Y} - Y) = (\widehat{Y} - AX)^\top \Sigma^{-1} (\widehat{Y} - AX) - 2(\widehat{Y} - AX)^\top \Sigma^{-1} \varepsilon + \varepsilon^\top \Sigma^{-1} \varepsilon.
\]
Taking the conditional expectation $\mathbb{E}_{Y|X}$:
\begin{align*}
\mathbb{E}_{Y|X} \left[ (\widehat{Y} - Y)^\top \Sigma^{-1} (\widehat{Y} - Y) \right] &= (\widehat{Y} - AX)^\top \Sigma^{-1} (\widehat{Y} - AX) - 0 + \mathbb{E}[\varepsilon^\top \Sigma^{-1} \varepsilon] \\
&= (\widehat{Y} - AX)^\top \Sigma^{-1} (\widehat{Y} - AX) + \operatorname{tr}(\Sigma^{-1} \Sigma) \\
&= (\widehat{Y} - AX)^\top \Sigma^{-1} (\widehat{Y} - AX) + n,
\end{align*}
where $n\in \mathbb{N}^{\star}$ is the dimension of $Y$.

Therefore,
\[
\mathbb{E}_{Y|X} \left[ -\frac{\lambda}{2} (\widehat{Y} - Y)^\top \Sigma^{-1} (\widehat{Y} - Y) \right] = -\frac{\lambda}{2} \left[ (\widehat{Y} - AX)^\top \Sigma^{-1} (\widehat{Y} - AX) + n \right].
\]

Now, the log-likelihood of $\widehat{Y}$ under $q(\cdot|X)$ is:
\[
\log q(\widehat{Y}|X) = -\frac{1}{2} (\widehat{Y} - AX)^\top \Sigma^{-1} (\widehat{Y} - AX) - \frac{1}{2} \log\left((2\pi)^n |\Sigma|\right).
\]
Thus,
\[
(\widehat{Y} - AX)^\top \Sigma^{-1} (\widehat{Y} - AX) = -2 \log q(\widehat{Y}|X) - \log\left((2\pi)^n |\Sigma|\right).
\]

one has:
\begin{align*}
\mathbb{E}_{Y|X} \left[ -\frac{\lambda}{2} (\widehat{Y} - Y)^\top \Sigma^{-1} (\widehat{Y} - Y) \right] &= -\frac{\lambda}{2} \left[ -2 \log q(\widehat{Y}|X) - \log\left((2\pi)^n |\Sigma|\right) + n \right] \\
&= \lambda \log q(\widehat{Y}|X) + \underbrace{\frac{\lambda}{2} \left[ \log\left((2\pi)^n |\Sigma|\right) - n \right]}_{:=c_n}.
\end{align*}

The term $c_n$ is constant with respect to $\widehat{Y}$ and $\theta$. Therefore,

\[
\mathbb{E}_{X \sim q} \left[ \mathbb{E}_{\widehat{Y} \sim \hat{p}_\theta(\cdot|X)} \left[ \mathbb{E}_{Y|X} \left[ -\frac{\lambda}{2} (\widehat{Y} - Y)^\top \Sigma^{-1} (\widehat{Y} - Y) \right] \right] \right] = \lambda \mathbb{E}_{X \sim q} \left[ \mathbb{E}_{\widehat{Y} \sim \hat{p}_\theta(\cdot|X)} \left[ \log q(\widehat{Y}|X) \right] \right] + c_n
\]

The entropy term is:
\[
\lambda \mathbb{H}(\hat{p}_\theta) = \lambda \mathbb{E}_{X \sim q} \left[ \mathbb{E}_{\widehat{Y} \sim \hat{p}_\theta(\cdot|X)} \left[ -\log \hat{p}_\theta(\widehat{Y}|X) \right] \right].
\]

Thus, the objective function becomes:
\begin{align*}
J(\theta) &= \lambda \mathbb{E}_{X \sim q} \left[ \mathbb{E}_{\widehat{Y} \sim \hat{p}_\theta(\cdot|X)} \left[ \log q(\widehat{Y}|X) - \log \hat{p}_\theta(\widehat{Y}|X) \right] \right] + c_n \\
&= -\lambda \mathbb{E}_{X \sim q} \left[ d_\text{KL} \left( \hat{p}_\theta(\cdot|X) \,\|\, q(\cdot|X) \right) \right] + c_n.
\end{align*}

So, maximizing $J(\theta)$ is equivalent to minimizing the reverse KL divergence, which completes the proof.
\end{proof}

\subsection{On the definition of $\mathcal{H}$ in \cref{ass:reward}}
\label{app:hilbert}

Let \[
\mathcal{H} := \left( L^{2}\!\left( \mathbb{R}^{n} \times \mathbb{R}^{n}, \, \mathbb{R}, \, e^{ -\| X - X' \|^{2} } \, d\lambda(X, X') \right); \, \langle \cdot, \cdot \rangle_{\mathcal{H}} \right),
\]

where
\[ \forall f,g \in \mathcal{H}\quad 
\langle f, g \rangle_{\mathcal{H}} = \int_{\mathbb{R}^{n}\times \mathbb{R}^{n}} f(X, X') \, g(X, X') \, e^{-\|X - X'\|^{2}} \, d\lambda(X) \, d\lambda(X')
\]

and \(d\lambda\) denotes the Lebesgue measure.
\begin{boxlem}
    Let $U\in \mathbb{R}^{n\times n}$, then $r_U$ as defined in \ref{ass:reward} is an element of \[
\mathcal{H} := L^{2}\!\left( \mathbb{R}^{n} \times \mathbb{R}^{n}, \, \mathbb{R}, \, e^{ -\| X - X' \|^{2} } \, d\lambda(X, X') \right).
\]
\end{boxlem}
\begin{proof}
We denote by $\langle \cdot, \cdot \rangle_{\mathbb{R}^n}$ the usual Euclidean scalar product on $\mathbb{R}^n$.  
In particular, for any $X, X' \in \mathbb{R}^n$ and any matrix $U \in \mathbb{R}^{n \times n}$, we have by the Cauchy-Schwarz inequality
\begin{equation}\label{C-S}
    \abs{(X - X')^{\top} U (X - X')} = \abs{\big\langle X - X', U(X - X') \big\rangle_{\mathbb{R}^n}}\leq \| X - X' \|^{2} \cdot \| U(X - X') \|^{2}
\end{equation}
and notice that for any $U \in \mathbb{R}^{n \times n}$ and $X \in \mathbb{R}^n$, 
\begin{equation}\label{ineqU}
    \| UX \|^{2} = \sum_{1 \leq k \leq n} \left( \sum_{j=1}^n U_{k,j} X_j \right)^{2} 
    \leq \underbrace{n^2 \left( \max_{(k,j) \in [1,n]^{2}} \abs{U_{k,j}}^{2} \right)}_{:= C(n,U) > 0} \|X \|^{2}.
\end{equation}

It leads to:

\begin{align*}
&\int_{\mathbb{R}^{n} \times \mathbb{R}^{n}}
    \big( - (X - X')^{\top} U (X - X') \big)^{2}
    e^{ -\| X - X' \|^{2} }
    \, d\lambda(X, X') 
    \\ &\underbrace{\leq}_{\text{\eqref{C-S}}} 
    \int_{\mathbb{R}^{n} \times \mathbb{R}^{n}}
    \| X - X' \|^{2} \cdot \| U(X - X') \|^{2}
    e^{ -\| X - X' \|^{2} }
    \, d\lambda(X, X') \notag \\
    &\underbrace{\leq}_{\eqref{ineqU}} 
    C(n,U) \int_{\mathbb{R}^{n} \times \mathbb{R}^{n}}
    \| X - X' \|^{4} 
    e^{ -\| X - X' \|^{2} }
    \, d\lambda(X, X') \notag \\
    &< \infty. \notag
\end{align*}
Since;
\begin{align*}
\int_{\mathbb{R}^n \times \mathbb{R}^n} \|X - X'\|^4 e^{-\|X - X'\|^2} d\lambda(X,X')
&= \int_{\mathbb{R}^n} \|Z\|^4 e^{-\|Z\|^2} dZ \\
&= \text{Vol}(S^{n-1}) \int_0^\infty r^{n-1} \cdot r^4 e^{-r^2} dr <\infty.
\end{align*}
 $$r_U\in \mathcal{H}.$$
\end{proof}

The following two lemmas justifies the reparametrization search space by $S_{n}^{++}(\mathbb{R})$.
\begin{boxlem}\label{phi_bijection}
The set $\left\{r_{U}\in \mathcal{H}\quad: U\in S^{++}_{n}(\mathbb{R}) \right\}$ is in bijection with $S_{n}^{++}(\mathbb{R}).$
\end{boxlem} 


\begin{proof}
    Denote $I:=\left\{r_{U}\in \mathcal{H}\quad: U\in S^{++}_{n}(\mathbb{R}) \right\}$ and 
    $$\varphi : S_{n}^{++}\to I$$ defined by 
    $$\forall U\in S_{n}^{++}\quad \varphi(U)=r_{U}.$$
    The surjectivity of $\varphi$ is straitghforward since the image of $\varphi$ is $I$. 
    Let $U_1,U_2\in S_{n}^{++}$ and assume $\phi(U_1)=\phi(U_2)$, which is 
    \begin{equation}\label{injectivity}
        \forall X, Y \in \mathbb{R}^n, \quad (X - Y)^T (U_1 - U_2) (X - Y) = 0.
    \end{equation}

But one can find $P\in GL_{n}(\mathbb{R})$ such that 
    \[
U_1 - U_2 = P D P^T, \quad D = \text{diag}(\lambda_1, \ldots, \lambda_n).
\]
Then \eqref{injectivity} reads:
\[\forall W=(w_1,...,w_n) \in \mathbb{R}^n \quad 
\sum_{i=1}^n \underbrace{\lambda_i w_i^2}_{\geq 0} = 0.
\]
Thus $\forall i\in [1,n]\quad \lambda_i=0$, so $U_1=U_2$.
\end{proof}

\begin{boxlem}
\label{lem:bijection_optim}
Let \( X \) and \( Y \) be two non-empty sets and let \( \phi : X \to Y \) be a bijection. Let \( f : X \to \mathbb{R} \) and \( g : Y \to \mathbb{R} \),
such that 
\begin{equation}\label{bij-eq}
    \forall x \in X \quad g(\phi(x))=f(x)
\end{equation}

Then the optimization problems:
\[
\text{(P1)} \quad \max_{x \in X} f(x) \quad \text{and} \quad \text{(P2)} \quad \max_{y \in Y} g(y),
\]
are equivalent in the sense that:
\begin{itemize}
    \item If \( x^* \) is a solution of (P1), then \( y^* = \phi(x^*) \) is a solution of (P2).
    \item Conversely, if \( y^* \) is a solution of (P2), then \( x^* = \phi^{-1}(y^*) \) is a solution of (P1).
\end{itemize}

\end{boxlem}

\begin{proof}
First; let's show that \eqref{bij-eq} implies that 
\begin{equation}
    \forall y\in Y\quad f(\phi^{-1}(y))=g(y).
\end{equation}

Let $y\in Y$, since $\phi$ is a bijection one can find $x\in X$ such that $y=\phi(x)$ and $x=\phi^{-1}(y)$ so using \eqref{bij-eq}: 
\begin{align}
    f \circ \phi^{-1}(y) 
        &= f(x)
        \underbrace{=}_{\eqref{bij-eq}}
        g \circ \phi(x)
        \underbrace{=}_{\text{bijectivity of }\phi}
        g \circ \phi \circ \phi^{-1}(y)
        = g(y).
\end{align}

Let's now proove that the sup of the two problems are equals. 

Since \( \phi \) is a bijection, every element \( y \in Y \) can be uniquely written as \( y = \phi(x) \) for some \( x \in X \). By assumption, we then have \( g(y) = g(\phi(x)) = f(x)\leq \sup_{x\in X}f(x):=M_f \). 

So \begin{equation}\label{eq:one}
    \sup_{y\in Y}g(y):=M_g\leq M_f
\end{equation}
Let $x\in X$, by the bijection, one can find $y\in Y$ such that $x=\phi^{-1}(y)$ so 
$f(x)=f(\phi^{-1}(y))=g(y)\leq M_g$. 

It follows that 
\begin{equation}\label{eq:two}
M_f\leq M_g
\end{equation}
Combining \eqref{eq:one} and \eqref{eq:two}:
$$ \sup_{y\in Y}g(y)=\sup_{x\in X}f(x).$$

Furthermore, if \( x^* \) is a point where \( f \) attains its maximum, then for \( y^* = \phi(x^*) \), we have:
\[
g(y^*) = f(x^*) = \max_{x \in X} f(x) = \max_{y \in Y} g(y),
\]
so \( y^* \) is a solution of (P2). Conversely, if \( y^* \) is a point where \( g \) attains its maximum, then let \( x^* = \phi^{-1}(y^*) \). We have:
\[
f(x^*) = g(\phi(x^*)) = g(y^*) = \max_{y \in Y} g(y) = \max_{x \in X} f(x),
\]
so \( x^* \) is a solution of (P1).
\end{proof}

\begin{boxprop}
 
For each $U \in S_n^{++}(\mathbb{R})$, define
\[
    f(U) = \mathbb{E}_{X}\Exp{Y\mid X \sim q}{  \log \hat{p}_{\theta_U}(Y \mid X) }.
\]
For each function $r \in I$, where
\[
    I = \left\{ r_U : U \in S_n^{++}(\mathbb{R}) \right\},
    \qquad \text{with} \qquad
    r_U(\widehat{Y}, Y) = -(\widehat{Y} - Y)^\top U (\widehat{Y} - Y),
\]
define
\[
    g(r) = \mathbb{E}_{X}\mathbb{E}_{Y\mid X \sim q}\log \hat{p}_{\theta_r}(Y \mid X) .
\]
Then the optimization problems
\[
    \max_{U \in S_n^{++}(\mathbb{R})} f(U)
    \qquad \text{and} \qquad
    \max_{r \in I} g(r)
\]
are equivalent.
\end{boxprop}

\begin{proof}
It's a straightforward consequence of the lemma \ref{lem:bijection_optim} with
$X:=S_{n}^{++}(\mathbb{R})$ and $Y:=I $
and the map \( \phi : X \to Y \) defined by \( \phi(U) = r_U \), for which we now that, by Lemma~\ref{phi_bijection}, is a bijection.

\end{proof}

\section{Additional experiments}
\label{app:exp}

\subsection{Distribution comparison}

\begin{figure}[h]
 \centering
 \includegraphics[width=\textwidth]{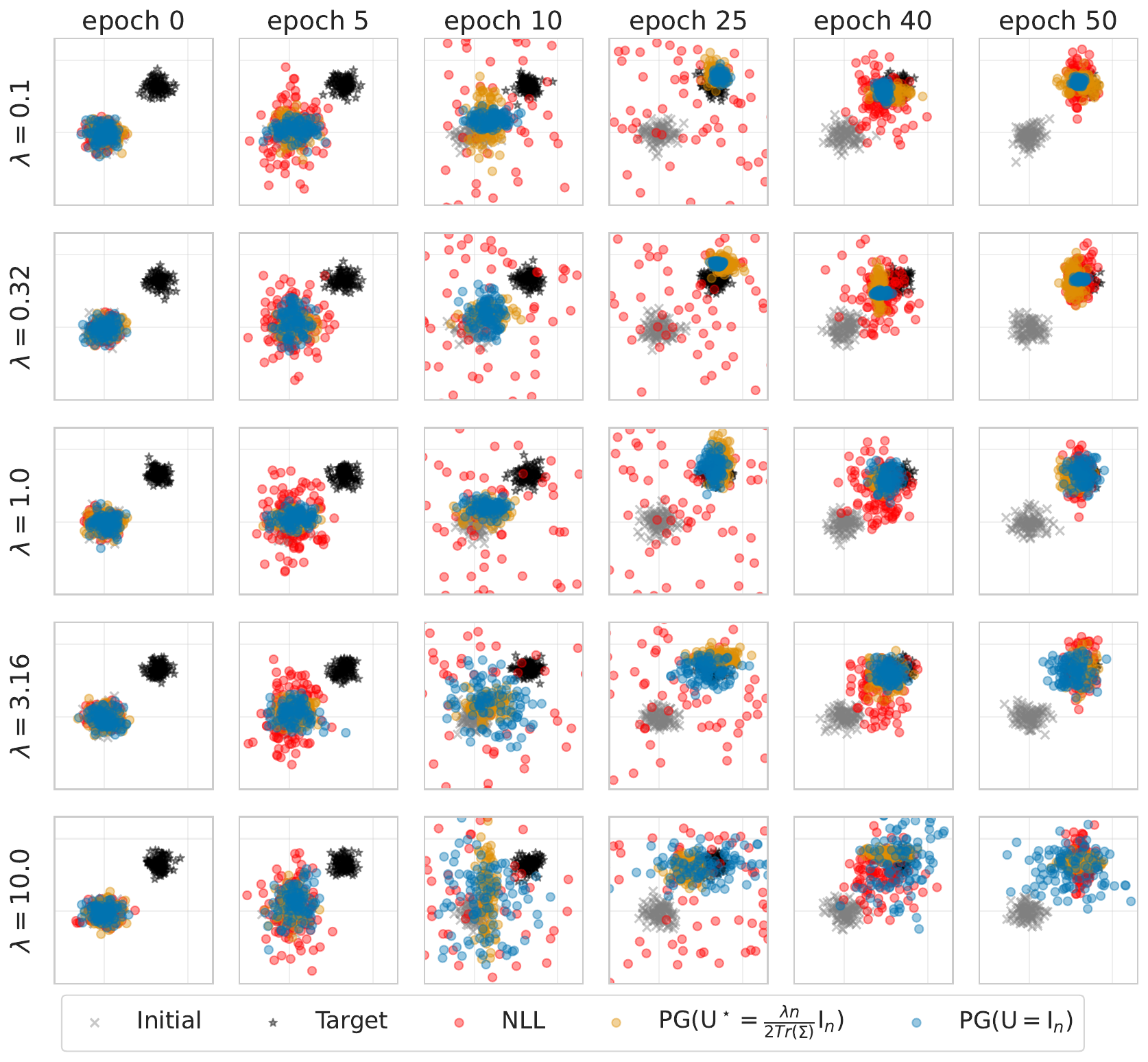}
 \caption{Distribution comparison, different value of $\lambda$}
 \label{fig:dist_app}
\end{figure}

\end{document}